\documentclass[10pt,twocolumn,letterpaper]{article}

\usepackage{cvpr}
\usepackage{times}
\usepackage{epsfig}
\usepackage{graphicx}
\usepackage{amsmath}
\usepackage{amssymb}
\usepackage{booktabs}
\usepackage{subcaption}
\usepackage{caption}
\usepackage{url}
\usepackage{xspace}
\usepackage{bbm}
\usepackage{dsfont}
\usepackage[numbers,sort,compress]{natbib}
\usepackage[font=small]{caption}

\usepackage{balance}

\usepackage{multirow}

\newcommand{\graspIt}{GraspIt\xspace}
\newcommand{\atlas}{AtlasNet\xspace}
\newcommand{\mano}{MANO\xspace}
\newcommand{\shapenet}{ShapeNet\xspace}
\newcommand{\fhb}{FHB\xspace}
\newcommand{\smpl}{SMPL\xspace}
\newcommand{\smplh}{SMPL+H\xspace}
\newcommand{\twoD}{2D\xspace}
\newcommand{\threeD}{3D\xspace}
\newcommand{\gt}{ground truth\xspace}

\newcommand{\datasetname}{ObMan }

 \usepackage[pagebackref=true,breaklinks=true,letterpaper=true,colorlinks,bookmarks=false]{hyperref}
\cvprfinalcopy 

\def\trimlen{1.0}

\begin{document}

\title{Learning joint reconstruction of hands and manipulated objects}

\author{
 Yana Hasson\textsuperscript{1,2}
 \qquad G\"{u}l Varol\textsuperscript{1,2}
 \qquad Dimitrios Tzionas\textsuperscript{3}
 \qquad Igor Kalevatykh\textsuperscript{1,2}\\
 \qquad Michael J. Black\textsuperscript{3}
 \qquad Ivan Laptev\textsuperscript{1,2}
 \qquad Cordelia Schmid\textsuperscript{1, 4}
 \\ \\
 {\normalsize \textsuperscript{1}Inria, \textsuperscript{2}D\'{e}partement d'informatique de l'ENS, CNRS, PSL Research University} \\
 {\normalsize	\textsuperscript{3}MPI for Intelligent Systems, T{\"u}bingen} \\
 {\normalsize	\textsuperscript{4}Univ. Grenoble Alpes, CNRS, Grenoble INP,
 LJK}}

\maketitle

\begin{abstract}

Estimating hand-object manipulations is essential for interpreting and imitating
human actions.  Previous work has made significant progress towards
reconstruction of hand poses and object shapes in isolation.  Yet,
reconstructing hands and objects during manipulation is a more challenging task
due to significant occlusions of both the hand and object.  While presenting
challenges, manipulations may also simplify the problem since the physics of
contact restricts the space of valid hand-object configurations. 
For example, during
manipulation, the hand and object should be in contact but not interpenetrate.
In this work, we regularize the joint reconstruction of
hands and objects with manipulation constraints. We present an end-to-end learnable model that exploits a
novel contact loss that favors physically plausible hand-object constellations.
Our approach
improves grasp quality metrics over baselines,
using RGB images as input.
To train and evaluate the model, we also propose a new large-scale synthetic
dataset, \mbox{ObMan}, with hand-object manipulations.
We demonstrate the transferability of ObMan-trained models to real data.

\end{abstract}

\mbox{}\vspace{-1cm}\\
\section{Introduction}
\label{sec:intro}

Accurate estimation of human hands,
as well as their interactions with the physical world, is vital to better
understand human actions and interactions.  In particular, recovering the
\threeD shape of a hand is key to many
applications including virtual and augmented reality, human-computer
interaction, action recognition and imitation-based learning of robotic skills.

\begin{figure}
\centering
\includegraphics[width=0.83\linewidth]{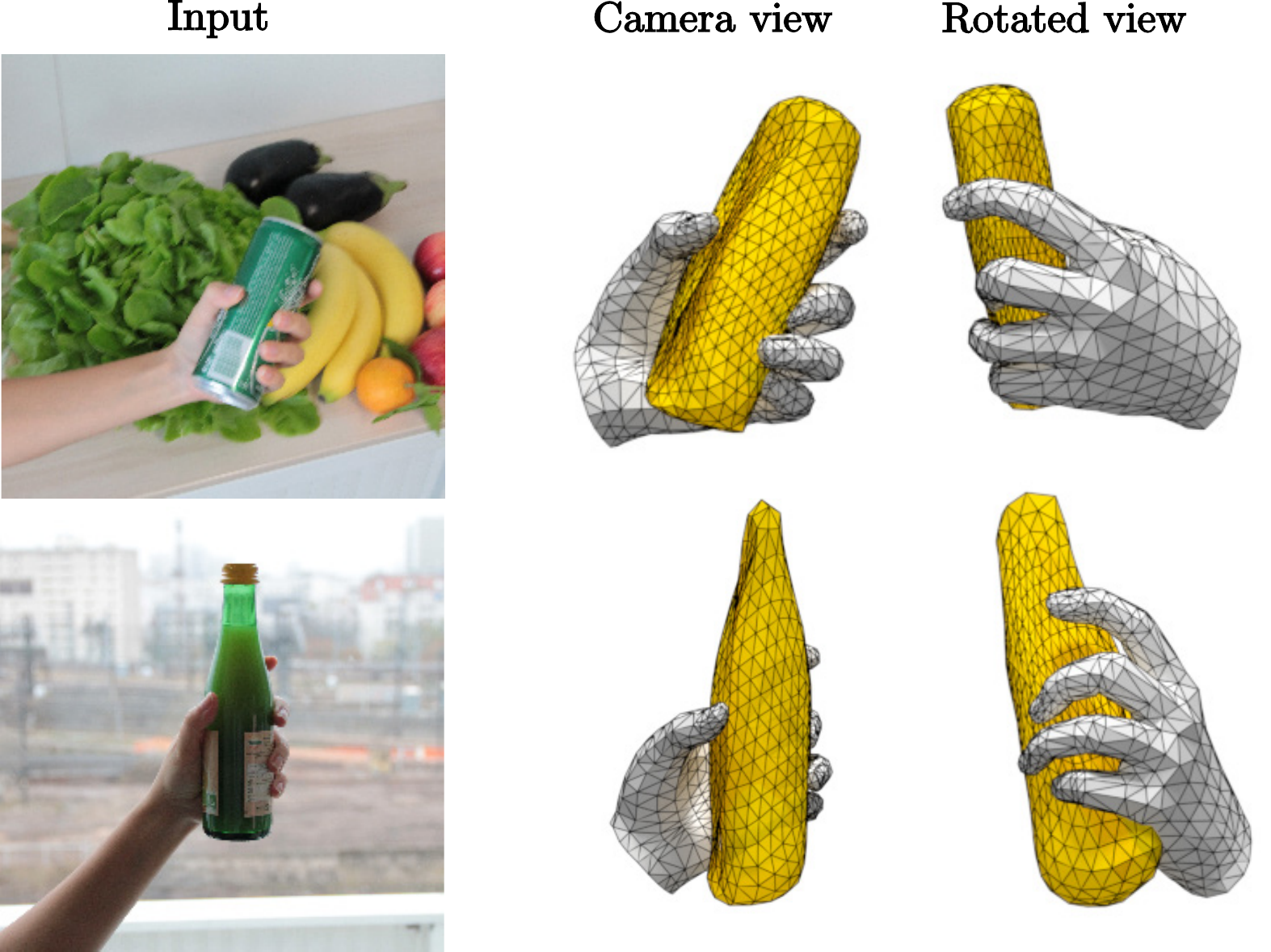}
\caption{Our method jointly reconstructs hand and object meshes from a monocular RGB
image.
Note that the model generating the predictions for the above images, which we
captured with an ordinary camera, was trained only on images from our synthetic
dataset, ObMan.}
\label{fig:teaser}
\mbox{}\vspace{-0.8cm}\\
\end{figure}

Hand analysis in images and videos has a long history in computer vision.
Early work focused on hand estimation and tracking using articulated models~\cite{Kanade94,HoggHand96,Wu2001,Cipolla_ModelBased_2001} or statistical shape models~\cite{Isard2000}.
The advent of RGB-D sensors brought remarkable progress to hand pose estimation from depth images \cite{Hamer_Hand_Manipulating_2009,OikonomidisBMVC_2011,KeskinECCV12,TKKIM_ICCV13_Real_time_Articulated_Hand,NYU_tracker_tompson14tog}.
While depth sensors provide strong cues, their applicability is limited by the energy consumption and environmental constrains such as distance to the target and exposure to sunlight.
Recent work obtains promising results for 2D and 3D hand pose estimation from
monocular RGB images using convolutional neural
networks~\cite{simon2017hand,brox:ICCV:2017,iqbal2018ECCV,GANeratedHands_CVPR2018,Dibra2018a,spurr2018cvpr,Panteleris2018}.
Most of this work, however, targets sparse keypoint
estimation which is not sufficient for reasoning about hand-object contact.
Full 3D hand meshes are sometimes estimated from images
by fitting a hand mesh to detected joints \cite{Panteleris2018} or by tracking given a good initialization \cite{ParagiosHandMonocular2011}.
Recently, the 3D {\em shape} or {\em surface} of a hand using an end-to-end
learnable model has been addressed with depth input~\cite{hps2018}.

\begin{figure}
   \centering
   \includegraphics[width=\linewidth]{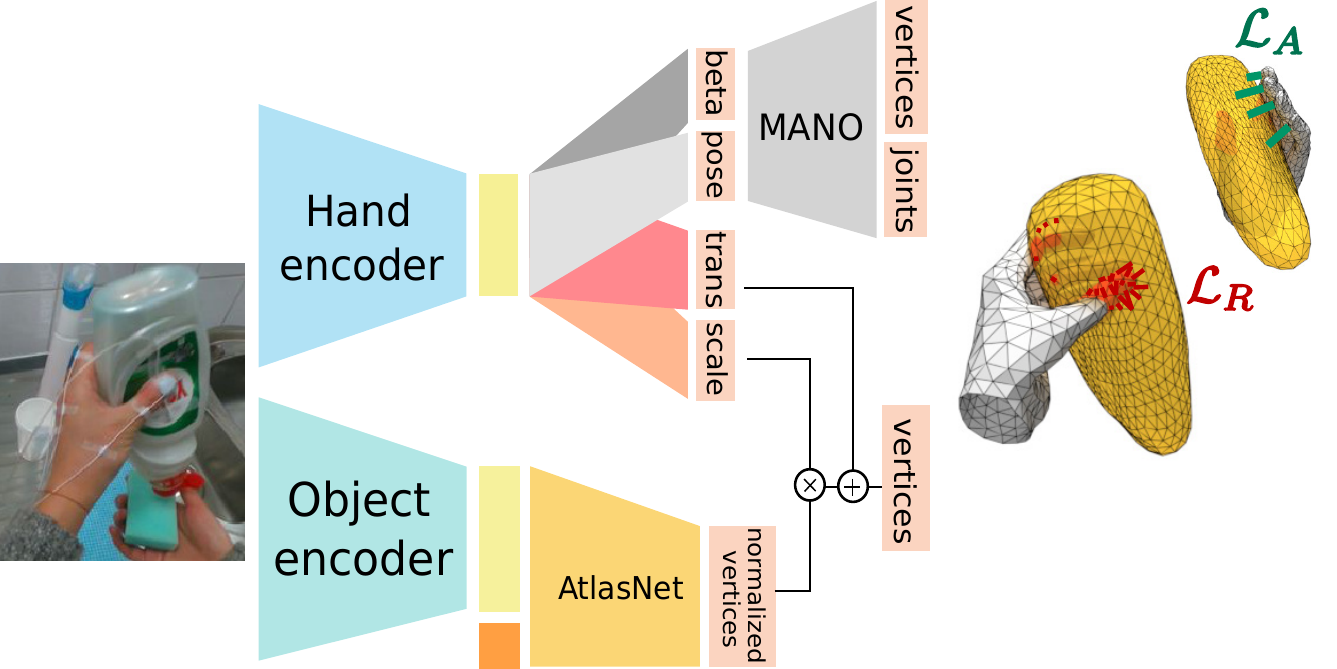}
   \caption{Our model predicts the hand and object meshes in a single forward
   pass in an end-to-end framework. The repulsion loss $\mathcal{L}_{R}$ penalizes interpenetration
   while the attraction loss $\mathcal{L}_{A}$ encourages the contact regions to be in contact with the object.}
   \label{fig:architecture}
   \mbox{}\vspace{-0.8cm}\\
\end{figure}

Interactions impose constraints on relative configurations of hands and objects.
For example, stable object grasps require contacts between hand and object
surfaces, while solid objects prohibit penetration. In this work we exploit
constraints imposed by object manipulations to reconstruct hands and objects as
well as to model their interactions. We build on a parametric hand model,  \mano
\cite{MANO:SIGGRAPHASIA:2017}, derived from 3D scans of human hands, that
provides anthropomorphically valid hand meshes. We then propose a
differentiable \mano network layer enabling end-to-end learning of hand shape
estimation. Equipped with the differentiable shape-based hand model, we next design a
network architecture for joint estimation of hand shapes, object shapes and
their relative scale and translation.  We also propose a novel contact loss that
penalizes penetrations and encourages contact between hands and manipulated
objects. An overview of our method is illustrated in
Figure~\ref{fig:architecture}.

Real images with ground truth shape for interacting hands and objects are difficult to obtain in practice.
Existing datasets with hand-object interactions are either too small for
training deep neural networks~\cite{Tzionas:IJCV:2016} or provide only partial \threeD
hand or object annotations~\cite{RealtimeHO_ECCV2016}.
The recent dataset by Garcia-Hernando~\etal~\cite{hernando2018cvpr} provides
\threeD hand joints and meshes of $4$ objects during hand-object interactions.

Synthetic datasets are an attractive alternative given their scale and
readily-available ground truth. Datasets with synthesized hands have been
recently introduced~\cite{brox:ICCV:2017,GANeratedHands_CVPR2018,hps2018} but
they do not contain hand-object interactions.
We generate a new large-scale synthetic dataset with objects manipulated by
hands: ObMan (\textit{Ob}ject \textit{Man}ipulation).  We achieve diversity by automatically generating hand grasp poses
for $2.7K$ everyday object models from $8$ object categories.
We adapt \mano to an automatic grasp generation tool
based on the \graspIt software~\cite{Miller2004}.
ObMan is sufficiently large and diverse to support training and ablation studies
of our deep models, and sufficiently realistic to generalize to real
images. See Figure~\ref{fig:teaser} for reconstructions obtained for real images
when training our model on ObMan.

In summary we make the following contributions.
First, we design the first end-to-end learnable model for joint 3D
reconstruction of hands and objects from RGB data.  Second, we propose a novel
contact loss penalizing penetrations and encouraging contact between hands and
objects.  Third, we create a new large-scale synthetic dataset, ObMan, with
hand-object manipulations. The ObMan dataset and our pre-trained models and code
are publicly available\footnote{\url{http://www.di.ens.fr/willow/research/obman/}}.

\section{Related work}
\label{sec:relatedwork}

In the following, we 
review methods that address hand and object
reconstructions in isolation. We then present
related works that
jointly reconstruct hand-object interactions.

\noindent\textbf{Hand pose estimation.}
Hand pose estimation
has attracted a lot of research interest since the $90s$ \cite{HoggHand96,Kanade94}.
The availability of commodity RGB-D sensors~\cite{kinectWiki,primesenseWiki,shottonCVPR2011}
led to significant progress in estimating \threeD hand pose given depth
or RGB-D
input~\cite{Hamer_Hand_Manipulating_2009,KeskinECCV12,Lepetit:ICCV:2015:handCnnLoop,oberweger2015handsDeep}.
Recently, the community has shifted its focus to RGB-based methods
\cite{iqbal2018ECCV,GANeratedHands_CVPR2018,Panteleris2018,simon2017hand,brox:ICCV:2017}.
To overcome the lack of \threeD annotated data,
many methods employed synthetic training
images~\cite{Dibra2018a,GANeratedHands_CVPR2018,Mueller2017ICCV,brox:ICCV:2017,hps2018}.
Similar to these approaches, we make use of synthetic
renderings, but we additionally integrate object
interactions. 

\threeD hand pose estimation has often been treated as predicting
\threeD positions of \textit{sparse} joints
\cite{iqbal2018ECCV,GANeratedHands_CVPR2018,brox:ICCV:2017}.
Unlike methods that predict only skeletons, our focus is to
output a \textit{dense} hand mesh to be able to infer
interactions with objects.
Very recently, Panteleris~\etal~\cite{Panteleris2018}
and Malik~\etal~\cite{hps2018} produce full hand meshes.
However, \cite{Panteleris2018} achieves this as a post-processing
step by fitting to \twoD predictions. Our hand estimation
component is most similar to \cite{hps2018}.
In contrast to \cite{hps2018}, our
method takes not depth but RGB images as input, which is more challenging and
more general.

Regarding hand pose estimation in the presence of objects, 
Mueller~\etal~\cite{Mueller2017ICCV,GANeratedHands_CVPR2018} grasp 7 objects
in a merged reality environment to render
synthetic hand pose datasets. However, objects only serve
the role of occluders, and the approach is difficult
to scale to more object instances.

\noindent\textbf{Object reconstruction.}
How to represent 3D objects in a CNN framework is an active research area.
Voxels~\cite{Maturana2015iros,marrnet}, point clouds~\cite{PointSet2017}, and
mesh surfaces~\cite{groueix2018,kato2018renderer,wang2018pixel2mesh} have been
explored. 
We employ the latter since meshes allow better modeling of the interaction with the hand.
\atlas~\cite{groueix2018} inputs vertex coordinates concatenated
with image features and outputs a deformed mesh. More recently,
Pixel2Mesh~\cite{wang2018pixel2mesh} explores regularizations
to improve the perceptual quality of predicted meshes.
Previous works mostly focus on producing accurate
shape and they output the object in a normalized coordinate
frame in a category-specific canonical pose. We employ a view-centered variant
of~\cite{groueix2018} to handle generic
object categories, without any category-specific knowledge.
Unlike existing methods that
typically input simple renderings
of CAD models, such as ShapeNet~\cite{ShapeNet2015},
we work with complex images in the presence of hand occlusions. 
In-hand scanning~\cite{RusinkiewiczRealTimeINHAND,weise_inHand_CVIU11,Tzionas:ICCV:2015,Argyros:BMVC:2015},
while performed in the context of manipulation, focuses on object reconstruction
and requires RGB-D video inputs. 

\noindent\textbf{Hand-object reconstruction.}
Joint reconstruction of hands and objects has been studied
with multi-view RGB \cite{Oikonomidis_1hand_object,LucaHands,wang2013handObject}
and RGB-D input with either optimization
\cite{Hamer_Hand_Manipulating_2009,Hamer_ObjectPrior,Oikonomidis_2hands,Tzionas:IJCV:2016,Tzionas:ICCV:2015,RealtimeHO_ECCV2016,Pham_2018_TPAMI,Tsoli2018} 
or classification \cite{romero2010b,Rogez:ECCVw:2014,Rogez:CVPR:2015,RogezSR15} 
approaches. 
These works use rigid objects, except for a few that use articulated \cite{Tzionas:IJCV:2016} 
or deformable objects \cite{Tsoli2018}.
Focusing on contact points, most works employ proximity
metrics \cite{RealtimeHO_ECCV2016,Tzionas:IJCV:2016,Tsoli2018},
while \cite{RogezSR15} directly
regresses them from images, 
and \cite{Pham_2018_TPAMI} uses contact measurements on instrumented objects.
\cite{Tzionas:IJCV:2016} integrates physical constraints for penetration and
contact, attracting fingers onto the object uni-directionally. 
On the contrary, \cite{Tsoli2018} symmetrically attracts the fingertips and the object surface. 
The last two approaches evaluate all possible configurations of contact points 
and select the one that provides the most stable grasp~\cite{Tzionas:IJCV:2016} or best matches visual evidence~\cite{Tsoli2018}. 
Most related to our work, given an RGB image, Romero~\etal~\cite{romero2010b}
query a large synthetic dataset of rendered hands interacting with objects to retrieve
configurations that match the visual evidence.
Their method's accuracy, however, is limited by the variety of configurations
contained in the database.
In parallel work to ours \cite{tekin19_handplusobject} jointly estimates hand
skeletons and 6DOF for objects.

Our work differs from previous hand-object reconstruction
methods mainly by incorporating
an end-to-end learnable CNN architecture
that benefits from a differentiable hand model and
differentiable physical constraints on penetration and contact.

\section{Hand-object reconstruction}
\label{sec:method}

As illustrated in Figure~\ref{fig:architecture},
we design a neural network architecture
that reconstructs the hand-object configuration in a single forward
pass from a rough image crop of a left hand holding an object.
Our network architecture is split into two branches.
The first branch reconstructs the object shape in a normalized coordinate space.
The second branch predicts the hand mesh as well as the information necessary to
transfer the object to the hand-relative coordinate system.
Each branch has a ResNet18~\cite{He2015} encoder
pre-trained on ImageNet~\cite{ILSVRC15}.
At test time, our model can process 20fps on a Titan X GPU.
In the following, we detail the three
components of our method: hand mesh estimation
in Section~\ref{subsec:hand}, object mesh estimation
in Section~\ref{subsec:object}, and the contact between
the two meshes in Section~\ref{subsec:contact}.

\subsection{Differentiable hand model}
\label{subsec:hand}

Following the methods that integrate the \smpl parametric body
model~\cite{SMPL:2015} as a network
layer~\cite{hmrKanazawa17,pavlakos2018humanshape}, we integrate the \mano hand
model~\cite{MANO:SIGGRAPHASIA:2017} as a differentiable layer.
\mano is a
statistical model that maps pose ($\theta$) and shape ($\beta$) parameters to a mesh.
While the pose parameters capture the angles between hand joints, the shape
parameters control the person-specific deformations of the hand;
see~\cite{MANO:SIGGRAPHASIA:2017} for more details.

Hand pose lives in a low-dimensional
subspace~\cite{MANO:SIGGRAPHASIA:2017,Lin:2000:MCH:822088.823446}.
Instead of predicting the full $45$-dimensional pose space, we predict $30$ pose
PCA components.
We found that performance saturates at $30$ PCA components and keep this value
for all our experiments (see Appendix~\ref{app:subsec:manorepresentation}).

\noindent\textbf{Supervision on vertex and joint positions 
($\mathcal{L}_{V_{\mathit{Hand}}}, \mathcal{L}_{J}$).}
The hand encoder produces an encoding $\Phi_{\mathit{Hand}}$ from an image.
Given $\Phi_{\mathit{Hand}}$, a fully connected network regresses $\theta$ and $\beta$.
We integrate the mesh generation as a differentiable network layer
that takes $\theta$ and $\beta$ as inputs and outputs the hand vertices
$V_{\mathit{Hand}}$ and $16$ hand joints.
In addition to \mano joints, we select $5$ vertices on the mesh as fingertips to
obtain $21$ hand keypoints $J$.
We define the supervision on the vertex positions
($\mathcal{L}_{V_{\mathit{Hand}}}$) and joint
positions ($\mathcal{L}_{J}$) to enable training on datasets where a ground truth
hand surface is not available.
Both losses are defined as the L2 distance to the ground truth.
We use root-relative \threeD positions as supervision for
$\mathcal{L}_{V_{\mathit{Hand}}}$ and $\mathcal{L}_{J}$.
Unless otherwise specified, we use the wrist defined by \mano as the root joint.

\noindent\textbf{Regularization on hand shape ($\mathcal{L}_{\beta}$).}
Sparse supervision can cause extreme mesh deformations when the hand shape is
unconstrained.
We therefore use a regularizer, $\mathcal{L}_{\beta}~=~\|\beta\|^2$, on the hand
shape to constrain it to be close to the average shape in the \mano training
set, which corresponds to $\beta=\vec{0} \in \mathds{R}^{10}$.

The resulting hand reconstruction loss $\mathcal{L}_{\mathit{Hand}}$ is the
summation of all $\mathcal{L}_{V_{\mathit{Hand}}}$, $\mathcal{L}_{J}$
and $\mathcal{L}_{\beta}$ terms:
\begin{equation}
    \mathcal{L}_{\mathit{Hand}} = \mathcal{L}_{V_{\mathit{Hand}}} +
    \mathcal{L}_{J} + \mathcal{L}_{\beta} .
\end{equation}
Our experiments indicate benefits
for all three terms (see Appendix~\ref{app:subsec:manoloss}). Our hand branch also matches
state-of-the-art performance on a standard benchmark for \threeD hand pose
estimation (see Appendix~\ref{app:subsec:handsoa}).
\subsection{Object mesh estimation}
\label{subsec:object}

Following recent methods~\cite{kato2018renderer,wang2018pixel2mesh},
we focus on genus 0 topologies.
We use \atlas~\cite{groueix2018} as the object
prediction component of our neural network architecture.
\atlas takes as input the concatenation of point coordinates sampled either on a set of
square patches or on a sphere, and
image features $\Phi_{\mathit{Obj}}$.
It uses a fully connected network to
output new coordinates on the surface of the reconstructed object.
\atlas explores two sampling strategies: sampling points from a sphere and sampling
points from a set of squares.
Preliminary experiments showed better generalization to unseen classes
when input points were sampled on a sphere.
In all our experiments we deform an icosphere of subdivision level 3 which has
642 vertices.
\atlas was initially designed to reconstruct meshes in a canonical view.
In our model, meshes are reconstructed in
view-centered coordinates.
We experimentally verified that \atlas can accurately reconstruct meshes in this
setting (see Appendix~\ref{app:subsec:objectcanonical}).
Following \atlas, the supervision for object vertices is defined by the symmetric Chamfer loss between
the predicted vertices and points randomly sampled on the \gt external surface of
the object.

\noindent\textbf{Regularization on object shape ($\mathcal{L}_{E}, \mathcal{L}_{L}$).}
In order to reason about the inside and outside of the object, it is important
to predict meshes with well-defined surfaces and good quality triangulations.
However \atlas does not explicitly enforce constraints on mesh quality.
We find that when learning to model a limited number of object shapes,
the triangulation quality is preserved. However, when training
on the larger variety of objects of ObMan, we find
additional regularization on the object meshes beneficial.
Following \cite{wang2018pixel2mesh,cmrKanazawa18,groueix2018b} we
employ two losses that penalize irregular meshes.
We penalize edges with lengths different from the average edge length with an
edge-regularization loss, $\mathcal{L}_E$.
We further introduce a curvature-regularizing loss, $\mathcal{L}_{L}$, based on~\cite{cmrKanazawa18},
which encourages the curvature of the predicted mesh to be similar to the
curvature of a sphere (see details in
Appendix~\ref{app:subsec:objectregularization}.
We balance the weights of $\mathcal{L}_E$ and $\mathcal{L}_{L}$ by weights
$\mu_{E}$ and $\mu_{L}$ respectively, which we empirically set to 2 and 0.1.
These two losses together improve the quality of the predicted meshes, as we
show in Figure~\ref{fig:objectregularization} of the appendix.
Additionally, when training on the ObMan dataset, we first train the network to
predict normalized objects, and then freeze the object encoder and the AtlasNet decoder
while training the hand-relative part of the network. When training the objects in normalized coordinates, noted with $n$, the total object loss is:
\begin{equation}
    \mathcal{L}^{n}_{\mathit{Object}} = \mathcal{L}^n_{V_{\mathit{Obj}}} +  \mu_{L}
    \mathcal{L}_{L} + \mu_{E} \mathcal{L}_{E}.
\end{equation}

\noindent\textbf{Hand-relative coordinate system ($\mathcal{L}_{S},
\mathcal{L}_{T}$).} Following \atlas~\cite{groueix2018}, we first predict the
object in a normalized scale by offsetting and scaling the ground truth vertices
so that the object is inscribed in a sphere of fixed radius.
However, as we focus on hand-object interactions, we need to estimate the object position
and scale relative to the hand.
We therefore predict translation and scale in two branches, which
output the three offset coordinates for the translation (i.e., $x, y, z$) and a
scalar for the object scale.
We define
$\mathcal{L}_{T} = \Vert T - \hat{T} \Vert^2_2$ and
$\mathcal{L}_{S} = \Vert S - \hat{S} \Vert^2_2$,
where $\hat{T}$ and $\hat{S}$ are the predicted translation and scale. $T$ is the
ground truth object centroid in hand-relative coordinates and $S$ is the
ground truth maximum radius of the centroid-centered object.

\noindent\textbf{Supervision on object vertex positions ($\mathcal{L}^n_{V_{\mathit{Obj}}}, \mathcal{L}_{V_{\mathit{Obj}}} $).}
We multiply the \atlas decoded vertices
by the predicted scale and offset them according to the predicted
translation to obtain the final object reconstruction.
Chamfer loss ($\mathcal{L}_{V_{\mathit{Obj}}}$) is applied after translation and scale are applied.
When training in hand-relative coordinates the loss becomes:
\begin{equation}
    \mathcal{L}_{\mathit{Object}} = 
    \mathcal{L}_{T} + \mathcal{L}_{S} + 
    \mathcal{L}_{V_{\mathit{Obj}}}.
\end{equation}

\subsection{Contact loss}
\label{subsec:contact}

So far, the prediction of hands and objects does not leverage the
constraints that guide objects interacting in the physical world. Specifically,
it does not account for our prior knowledge that objects can not interpenetrate
each other and that, when grasping objects, contacts occur at the surface between
the object and the hand.
We formulate these contact constraints as a differentiable loss,
$\mathcal{L}_{Contact}$, which can be directly used in the end-to-end learning
framework.
We incorporate this additional loss using a weight parameter $\mu_{C}$, which we set
empirically to $10$.

We rely on the
following definition of distances between points.
$d(v, V_{\mathit{Obj}}) =
\inf_{w\in V_{\mathit{Obj}}} \Vert v - w \Vert_{2}$
denotes distances from point to set and
$d(C, V_{\mathit{Obj}}) = \inf_{v\in C}d(v, V_{\mathit{Obj}})$
denotes distances from set to set.
Moreover, we define a common penalization function $l_{\alpha}(x) = \alpha
\tanh\left(\frac{x}{\alpha}\right)$, where $\alpha$ is a characteristic
distance of action.

\noindent\textbf{Repulsion ($\mathcal{L}_{R}$).}
We define a repulsion loss ($\mathcal{L}_{R}$) that penalizes
hand and object \textit{interpenetration}.
To detect interpenetration, we first detect hand vertices that are inside the
object.
Since the object is a deformed sphere, it is watertight.
We therefore cast a ray from the hand vertex and count the number of times it
intersects the object mesh to determine whether it is inside or outside the
predicted mesh~\cite{Moller:1997:FMS:272313.272315}.
$\mathcal{L}_{R}$ affects all hand vertices that belong to
the interior of the object, which we denote $\mathrm{Int}(Obj)$.
The repulsion loss is defined as: 
\begin{equation*}
    \mathcal{L}_{R}(V_{\mathit{Obj}}, V_{\mathit{Hand}})  = \sum_{v \in V_{\mathit{Hand}}} \mathds{1}_{v \in
    \mathrm{Int}(V_{\mathit{Obj}})} l_r(d(v, V_{\mathit{Obj}})),
\end{equation*}
where $r$ is the repulsion characteristic distance, which we empirically set to
$2\mathrm{cm}$ in all experiments.

\noindent\textbf{Attraction ($\mathcal{L}_{A}$).}
We further define an attraction loss ($\mathcal{L}_{A}$) to penalize cases in
which hand vertices are in the vicinity of the object but the surfaces are
\textit{not} in contact. This loss is applied only to vertices which belong to
the exterior of the object $\mathrm{Ext}(Obj)$.

\begin{figure}
    \centering
    \includegraphics[width=0.35\linewidth]{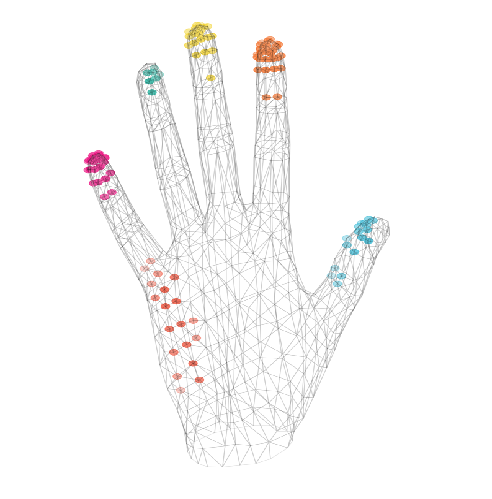}
    \includegraphics[width=0.59\linewidth]{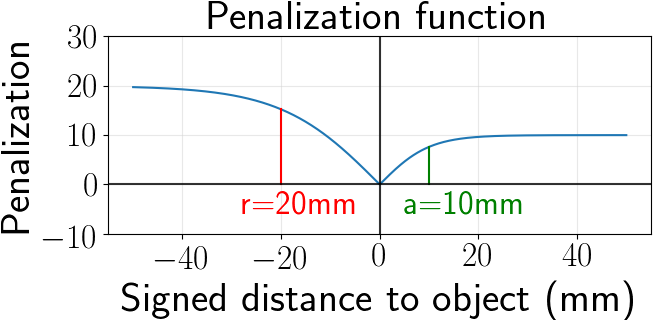}
    \caption{Left: Estimated contact regions from ObMan. We find that points
    that are often involved in contacts can be clustered into $6$ regions on
the palmar surface of the hand. Right: Generic shape of the penalization function emphasizing the role of the characteristic distances.}
    \label{fig:contact_zones}
    \mbox{}\vspace{-0.8cm}\\
\end{figure}

We compute statistics on the automatically-generated grasps
described in the next section to determine which
vertices on the hand are frequently involved in contacts.
We compute for each \mano vertex how often across the dataset it is in the
immediate vicinity of the object (defined as less than $3\textrm{mm}$ away from the object's
surface).
We find that by identifying the vertices that are close to the objects in at
least 8\% of the grasps, we obtain $6$ regions of connected vertices $\{{C_i}\}_{i 
\in [\![1, 6 ]\!]}$ on the hand which match the $5$ fingertips and part of
the palm of the hand, as illustrated in Figure~\ref{fig:contact_zones} (left).
The attraction term $\mathcal{L}_{A}$ penalizes distances from
each of the regions to the object, allowing for sparse guidance towards
the object's surface:
\begin{equation}
    \mathcal{L}_{A}(V_{\mathit{Obj}}, V_{\mathit{Hand}}) = \sum_{i=1}^{6}l_a(d(C_i \cap \mathrm{Ext}(Obj), V_{\mathit{Obj}})).
\end{equation}
We set $a$ to $1\mathrm{cm}$ in all experiments.
For regions that are further from the hand than a threshold $a$, the attraction will
significantly decrease and become negligible as the distance to the object
further increases, see Figure~\ref{fig:contact_zones} (right).

Our final contact loss $\mathcal{L}_{\mathit{Contact}}$ is a weighted sum of the attraction
$\mathcal{L}_{A}$  and the repulsion $\mathcal{L}_{R}$ terms:
\begin{equation}
    \mathcal{L}_{Contact} = \lambda_R \mathcal{L}_{R} + (1 - \lambda_R) \mathcal{L}_{A},
\end{equation}
where $\lambda_R \in [0, 1]$ is the contact weighting coefficient, e.g.,
$\lambda_R=1$ means only the repulsion term is active. We show in our experiments
that the balancing between attraction and repulsion is very important for
physical quality.

Our network is first trained with
$\mathcal{L}_{\mathit{Hand}}+\mathcal{L}_{\mathit{Object}}$.
We then continue training with 
$\mathcal{L}_{\mathit{Hand}}+\mathcal{L}_{\mathit{Object}}+\mu_C \mathcal{L}_{\mathit{Contact}}$
to improve the physical quality of the hand-object interaction. Appendix~\ref{app:subsec:training}
gives further implementation details.

\section{\datasetname dataset}
\label{sec:dataset}

To overcome the lack of adequate training data for our models, we generate a large-scale
synthetic image dataset of hands grasping objects which we call the \textit{ObMan}
dataset.
Here, we describe how we scale automatic generation of hand-object images.

\noindent\textbf{Objects.}
In order to find a variety of high-quality meshes of frequently manipulated
everyday objects, we selected models from the \shapenet~\cite{ShapeNet2015} dataset.
We selected $8$ object categories of everyday objects (bottles, bowls, cans,
jars, knifes, cellphones, cameras and remote controls).
This results in a total of $2772$ meshes which are split among the training,
validation and test sets.

\noindent\textbf{Grasps.} 
In order to generate plausible grasps, we use the \graspIt
software~\cite{Miller2004} following the methods used to collect the
Grasp Database~\cite{GraspDatabase2009}.
In the robotics community, this dataset has remained valuable over
many years~\cite{SahbaniEB12} 
and is still a reference for the fast
synthesis of grasps given known object
models~\cite{LenzLS13,mahler2017dex}.

We favor simplicity and robustness of the grasp
generation over the accuracy of the underlying model.
The software expects a rigid articulated model of the hand.
We transform \mano by separating it into 16 rigid parts, 3 parts for the
phalanges of each finger, and one for the hand palm.
Given an object mesh, \graspIt produces different grasps from various
initializations.
Following~\cite{GraspDatabase2009}, our generated grasps optimize for the grasp
metric but do not necessarily reflect the statistical distribution of human grasps.
We sort the obtained grasps according to a heuristic measure (see
Appendix~\ref{app:subsec:grasps})
and keep the two best candidates for each object. 
We generate a total of $21K$ grasps.

\begin{figure} 
\centerline{
\includegraphics[width=0.95\linewidth]{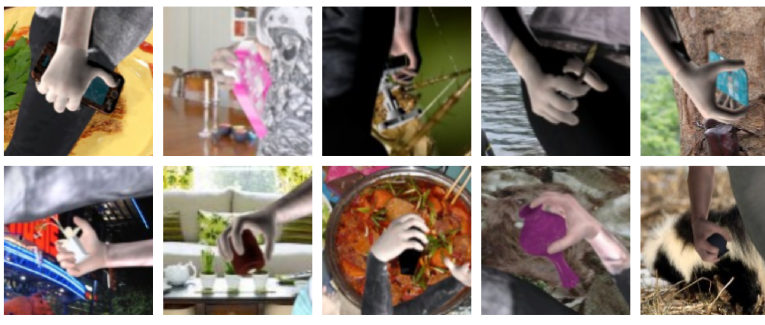}
}
\caption{{\bf ObMan:} large-scale synthetic dataset of hand-object interactions. We pose the \mano hand
model~\cite{MANO:SIGGRAPHASIA:2017}
to grasp~\cite{Miller2004} a given object mesh. The scenes are rendered 
with variation in texture, lighting, and background.}
\label{fig:dataset}
\mbox{}\vspace{-1cm}\\
\end{figure}

\noindent\textbf{Body pose.}
For realism, we render the hand and
the full body (see Figure \ref{fig:dataset}).
The pose of the hand is transferred to hands of the
\smplh~\cite{MANO:SIGGRAPHASIA:2017} model which integrates \mano to the
\smpl~\cite{SMPL:2015,MANO:SIGGRAPHASIA:2017} statistical body model, allowing
us to render realistic images of embodied hands.
Although we zoom our cameras to focus on the hands, we vary the body poses to provide
natural occlusions and coherent backgrounds.
Body poses and shapes are varied by sampling from the same distribution as in
SURREAL~\cite{varol17}; i.e., sampling poses from the CMU MoCap
database~\cite{cmu_mocap} and shapes from CAESAR~\cite{CAESAR}.
In order to maximize the viewpoint variability, a global rotation uniformly
sampled in $SO(3)$ is also applied to the body.
We translate the hand root joint to the camera's optical axis.
The distance to the camera is sampled uniformly between $50$ and $80\mathrm{cm}$.

\noindent\textbf{Textures.}
Object textures are randomly sampled from the texture maps provided with
\shapenet~\cite{ShapeNet2015} models.
The body textures are obtained from the full body scans used in
SURREAL~\cite{varol17}.  Most of the scans have missing color values in the hand
region.  We therefore combine the body textures with $176$ high resolution
textures obtained from hand scans from $20$ subjects.  The hand textures are
split so that textures from $14$ subjects are used for training and $3$ for
test and validation sets.
For each body texture, the skin tone of the hand is matched to the subject's face
color. Based on the face skin color, we query in the HSV color
space the $3$ closest hand texture matches.
We further shift the HSV channels of the hand to better match the person's skin
tone.

\noindent\textbf{Rendering.}
Background images are sampled from both the LSUN~\cite{yu_lsun} and
ImageNet~\cite{ILSVRC15} datasets.
We render the images using Blender~\cite{Blender2018}.
In order to ensure the hand and objects are visible we discard configurations
if less than $100$ pixels of the hand or if less than $40\%$ of
the object is visible.

For each hand-object configuration, we render object-only,
hand-only, and hand-object images, as well as the corresponding
segmentation and depth maps.

\section{Experiments} \label{sec:experiments}

We first define the evaluation metrics
and the datasets (Sections~\ref{subsec:metrics},~\ref{subsec:datasets})
for our experiments.
We then analyze the effects of occlusions (Section~\ref{subsec:occlusions})
and the contact loss (Section~\ref{subsec:effcontactloss}).
Finally, we present our transfer learning experiments from synthetic to real
domain (Sections~\ref{subsec:synth2real},~\ref{subsec:core50}).

\subsection{Evaluation metrics} \label{subsec:metrics}

Our output is structured, and a single metric does not fully capture
performance. We therefore rely on multiple evaluation metrics.

\noindent\textbf{Hand error.} For hand reconstruction, we compute the mean
end-point error (mm) over $21$ joints following~\cite{brox:ICCV:2017}.

\noindent\textbf{Object error.} Following \atlas~\cite{groueix2018},
we measure the accuracy of object reconstruction by
computing the symmetric
Chamfer distance (mm) between points sampled on the ground truth mesh and vertices of
the predicted mesh.

\noindent\textbf{Contact.} To measure the physical quality of our joint reconstruction,
we use the following metrics.

\textit{Penetration depth (mm), Intersection volume ($\textrm{cm}^3$):} 
Hands and objects should not share the same physical space.
To measure whether this rule is violated, we report the intersection volume
between the object and the hand as well as the penetration depth.
To measure the intersection volume of the hand and object we voxelize the hand and
object using a voxel size of $0.5\textrm{cm}$.
If the hand and the object collide, the penetration depth is the maximum
of the distances from hand mesh vertices to the object's surface.
In the absence of collision, the penetration depth is $0$.

\textit{Simulation displacement (mm):} Following \cite{Tzionas:IJCV:2016}, we use
physics simulation to evaluate the quality of the produced grasps. This metric
measures the average displacement of the object's center of mass in a simulated
environment~\cite{BulletPhysics} assuming the hand is fixed and the object is
subjected to gravity.
Details on the setup and the parameters used for the simulation can be found
in~\cite{Tzionas:IJCV:2016}.
Good grasps should be stable in simulation.
However, stable simulated grasps can also
occur if the forces resulting from the collisions balance each other.
For estimating grasp quality, simulated displacement must be
analyzed in conjunction with a measure of collision.
If both displacement in simulation and penetration depth are decreasing,
there is strong evidence that the physical quality of the grasp is improving
(see Section~\ref{subsec:effcontactloss} for an analysis).
The reported metrics are averaged across the dataset.

\subsection{Datasets} \label{subsec:datasets}

We present the datasets we use to evaluate our models.
Statistics for each dataset are summarized in Table~\ref{table:datasets}.

\noindent\textbf{First-person hand benchmark (\fhb).} This dataset~\cite{hernando2018cvpr} is a recent video
collection providing \threeD hand annotations for a wide range of hand-object interactions.
The joints are automatically annotated using magnetic sensors strapped on the 
hands, and which are visible on the RGB images.
\threeD mesh annotations are provided for four objects: three different bottles
and a salt box.
In order to ensure that the object being interacted with is unambiguously defined, we filter
frames in which the manipulating hand is further than $1\mathrm{cm}$ away
from the manipulated object. We refer to this filtered dataset as \fhb.
As the milk bottle is a genus-1 object and is often grasped by its handle, we
exclude this object from the experiments we conduct on contacts. We call this
subset FHB$_C$.
We use the same subject split as~\cite{hernando2018cvpr}, therefore, each object
is present in both the training and test splits.

The object annotations for this dataset suffer from some imprecisions.
To investigate the range of the object \gt error, we measure the penetration
depth of the hand skeleton in the object for each hand-object configuration.
We find that on the training split of $\mathrm{FHB}$, the average
penetration depth is $11.0\mathrm{mm}$ (std=$8.9\mathrm{mm}$).
While we still report quantitative results on objects for completeness, the \gt
errors prevent us from drawing strong conclusions from reconstruction metric
fluctuations on this dataset.

\noindent\textbf{Hands in action dataset (HIC).} We use a subset
of the HIC dataset~\cite{Tzionas:IJCV:2016} which has sequences
of a single hand interacting with objects.
This gives us $4$ sequences featuring manipulation of
a sphere and a cube.
We select the frames in which the hand is
less than $5\mathrm{mm}$ away from the object.
We split this dataset into $2$ training and $2$ test sequences with each object
appearing in both splits and restrict our predictions to the frames in which the minimal
distance between hand and object vertices is below $5\mathrm{mm}$.
For this dataset the hand and object meshes are provided.
We fit \mano to the provided hand mesh, allowing for
dense point supervision on both hands and objects.

\begin{table}
    \centering
    \resizebox{.99\linewidth}{!}{
        \begin{tabular}{lcccc}
            \toprule
            &  \datasetname & \fhb &  $\mathrm{FHB}_{C}$ & HIC \\
           \midrule
           \#frames & $141K/6K$  & $8420/9103$ & $5077/5657$ &
           $251/307$ \\
           \#video sequences  & - & $115/127$ & $76/88$  & $2/2$  \\
           \#object instances & $1947/411$ & $4$ & $3$ & $2$  \\
           real   & no &  yes & yes & yes  \\
            \bottomrule
        \end{tabular}
    }
    \caption{Dataset details for train/test splits.}
    \label{table:datasets}
\end{table}

\begin{table}
    \centering
    \resizebox{.99\linewidth}{!}{
        \begin{tabular}{lcc}
            \toprule
            & \multicolumn{2}{c}{Evaluation images} \\
            Training & H-img & HO-img \\
            \midrule
            H-img ($\mathcal{L}_{H}$) &  \textbf{10.3} & 14.1 \\
            HO-img ($\mathcal{L}_{H}$) & 11.7 & \textbf{11.6} \\
            \bottomrule
        \end{tabular}
        \begin{tabular}{lcc}
            \toprule
            & \multicolumn{2}{c}{Evaluation images} \\
            Training & O-img & HO-img \\
            \midrule
            O-img ($\mathcal{L}_{O}$) & \textbf{0.0242} &  0.0722 \\
            HO-img ($\mathcal{L}_{O}$) & 0.0319 & \textbf{0.0302}  \\
            \bottomrule
        \end{tabular}
    }
    \caption{We first show that training with occlusions is important when
    targeting images of hand-object interactions.}
    \label{table:occlusion}
    \mbox{}\vspace{-0.8cm}\\
\end{table}

\begin{figure*}
\centering
\includegraphics[width=0.9\linewidth]{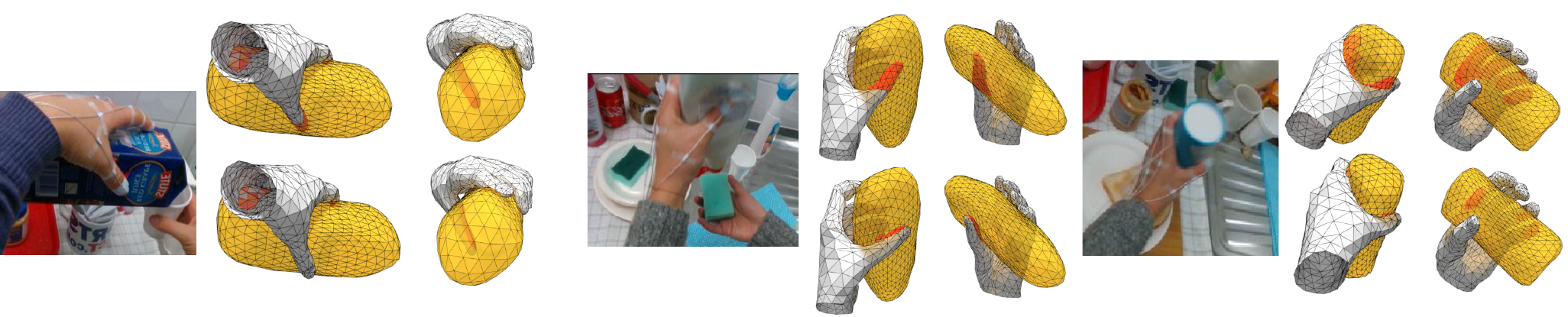}
\mbox{}\vspace{-0.4cm}\\
\caption{
    Qualitative comparison between \textit{with} (bottom) and \textit{without} (top) contact on
FHB$_C$. Note the improved contact and reduced penetration, highlighted with red
regions, with our contact loss.
}
\label{fig:contact_fhb_qual}
\mbox{}\vspace{-0.6cm}\\
\end{figure*}

\begin{table*}
    \centering
    \resizebox{.99\linewidth}{!}{
        \begin{tabular}{l|ccccc|ccccc}
            \toprule
            & \multicolumn{5}{c|}{\datasetname Dataset} &
            \multicolumn{5}{|c}{FHB$_C$ Dataset} \\
            & Hand & Object & Maximum & Simulation & Intersection &Hand & Object
            & Maximum & Simulation & Intersection \\
            & Error & Error & Penetration & Displacement & Volume & Error &
            Error & Penetration & Displacement & Volume \\
            \midrule
            No contact loss & 11.6 & 641.5 & 9.5 & 31.3 & 12.3 &
                            28.1 $\pm$ 0.5 & 1579.2 $\pm$ 66.2 & 18.7 $\pm$0.6
                                           & 51.2 $\pm$ 1.7 & 26.9 $\pm$ 0.2 \\
            Only attraction ($\lambda_{R}=0$) & 11.9 & 637.8 & 11.8 & 26.8 &
            17.4 &
            28.4 $\pm$ 0.6 & 1586.9 $\pm$ 58.3 & 22.7 $\pm$0.7 & 48.5 $\pm$ 3.2
                           & 41.2 $\pm$ 0.3 \\
            Only repulsion ($\lambda_{R}=1$) & 12.0  & 639.0  & 6.4 & 38.1 & 8.1 & 
                                             28.6 $\pm$ 0.8 & 1603.7 $\pm$ 49.9 & 6.0 $\pm$ 0.3 & 53.9 $\pm
                                             $ 2.3  & 7.1 $\pm$ 0.1 \\
            Attraction + Repulsion ($\lambda_{R}=0.5$) & 11.6 & 637.9 & 9.2 &
            30.9 & 12.2 &
            28.8 $\pm0.8$ & 1565.0 $\pm$ 65.9 & 12.1 $\pm$ 0.7 & 
            47.7 $\pm2.5$ & 17.6 $\pm$ 0.2  \\
            \bottomrule
        \end{tabular}
    }
    \caption{We experiment with each term of the contact loss.
  Attraction ($\mathcal{L}_{A}$) encourages contacts between close points while
  repulsion ($\mathcal{L}_{R}$) penalizes interpenetration. $\lambda_{R}$ is the
  repulsion weight, balancing the contribution of the two terms.  }
    \label{table:contact}
\mbox{}\vspace{-0.9cm}\\
\end{table*}

\begin{figure}
\includegraphics[width=0.465\linewidth]{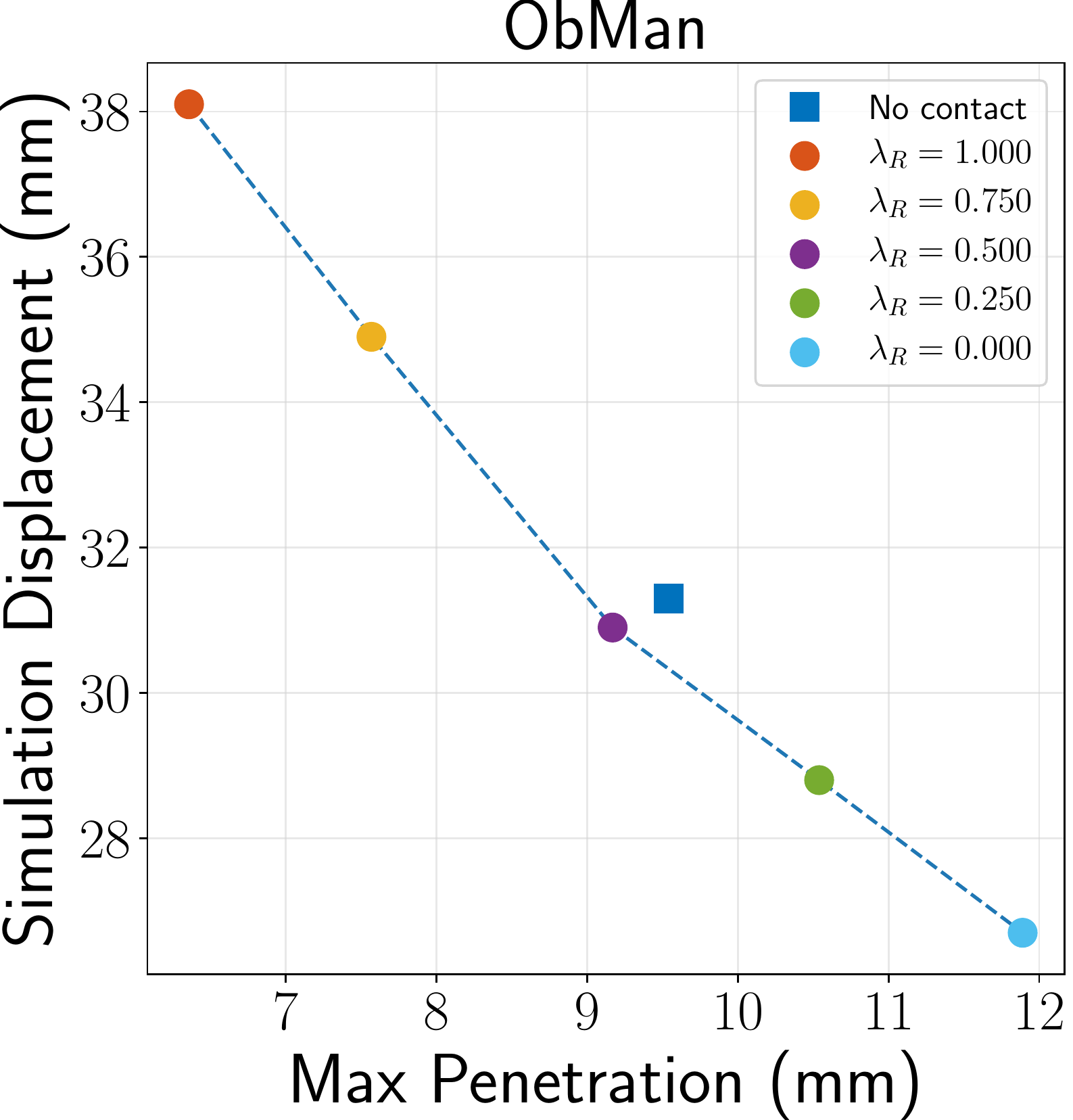}
\includegraphics[width=0.47\linewidth]{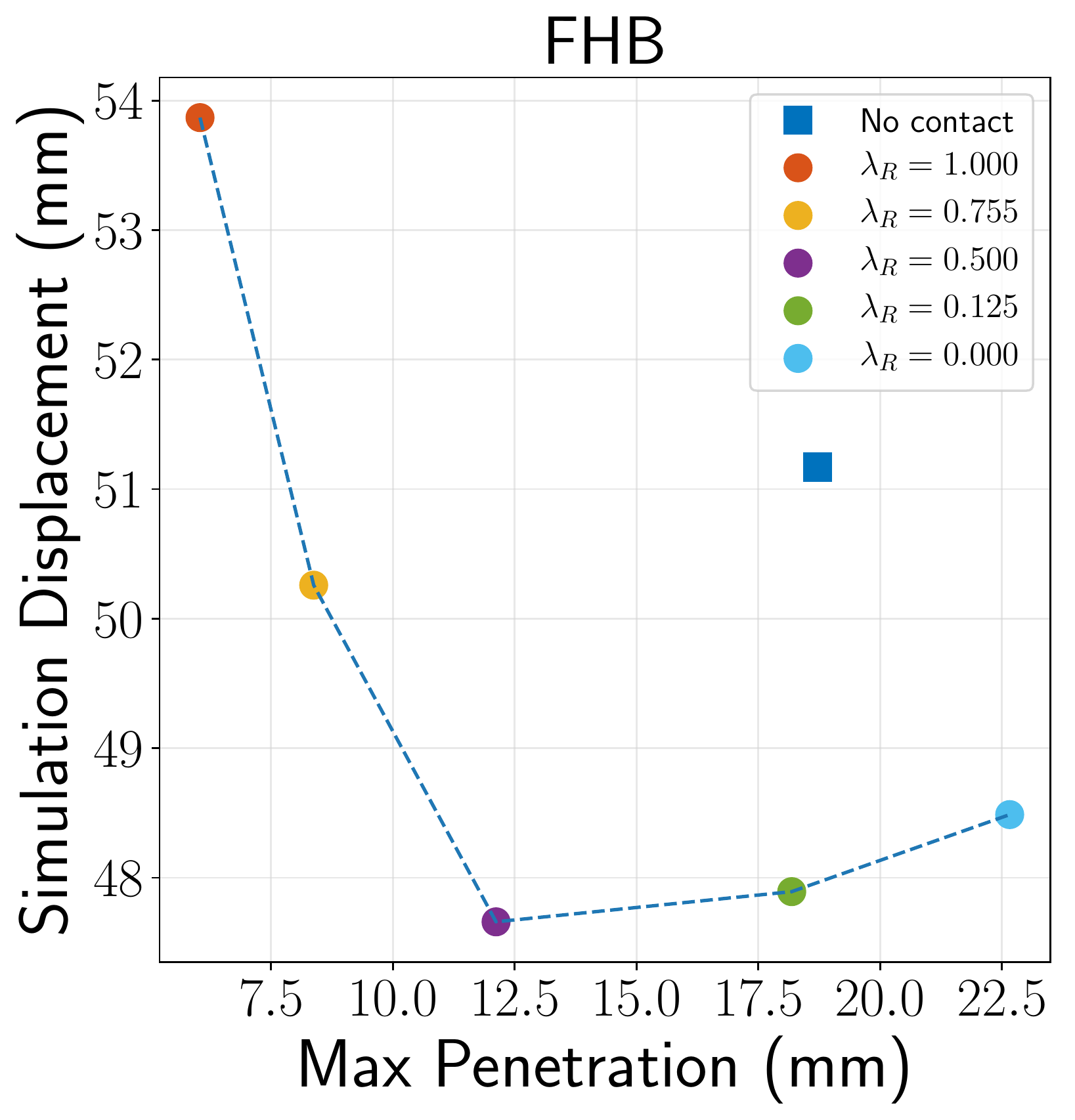}
\caption{We examine the relative importance between the contact terms
on the grasp quality metrics. Introducing a well-balanced
contact loss improves upon the baseline on both max penetration and simulation
displacement.}
\label{fig:contactcurve}
\mbox{}\vspace{-1cm}\\
\end{figure}

\subsection{Effect of occlusions} \label{subsec:occlusions}

For each sample in our synthetic dataset, in addition to the hand-object image (HO-img) we
render two images of the corresponding isolated and unoccluded hand (H-img) or
object (O-img).
With this setup, we can systematically study the effect of occlusions on ObMan,
which would be impractical outside of a synthetic setup.

We study the effect of objects occluding hands by training two networks, one trained on
hand-only images and one on hand-object images.
We report performance on both unoccluded and occluded images.
A symmetric setup is applied to study the effect of hand occlusions on objects.
Since the hand-relative coordinates are not applicable to experiments with
object-only images, we study the normalized shape reconstruction,
centered on the object centroid, and scaled to be inscribed in a sphere of
radius 1.

Unsurprisingly, the best performance is obtained when both training and testing
on unoccluded images as shown in Table~\ref{table:occlusion}.
When both training and testing on occluded images, reconstruction errors for
hands and objects drop significantly, by $12\%$ and $25\%$
respectively.  This validates the intuition that estimating hand pose and object
shape in the presence of occlusions is a harder task.

We observe that for both hands and objects, the most challenging setting 
is training on unoccluded images while testing on images with occlusions.
This shows that training with occlusions is crucial for accurate
reconstruction of hands-object configurations.

\subsection{Effect of contact loss} \label{subsec:effcontactloss}

In the absence of explicit physical constraints, the predicted hands and
objects have an average penetration depth of $9\mathrm{mm}$ for \datasetname
and $19\mathrm{mm}$ for $\mathrm{FHB}_C$ (see Table~\ref{table:contact}).
The presence of interpenetration at test time shows that the model is not
implicitly learning the physical rules governing hand-object manipulation.
The differences in physical metrics between the two datasets can be attributed
to the higher reconstruction accuracy for \datasetname but also to
the noisy object ground truth in $\mathrm{FHB}_C$ which
produces penetrated and likely unstable `\gt' grasps.

\begin{figure*}
\includegraphics[width=0.99\linewidth]{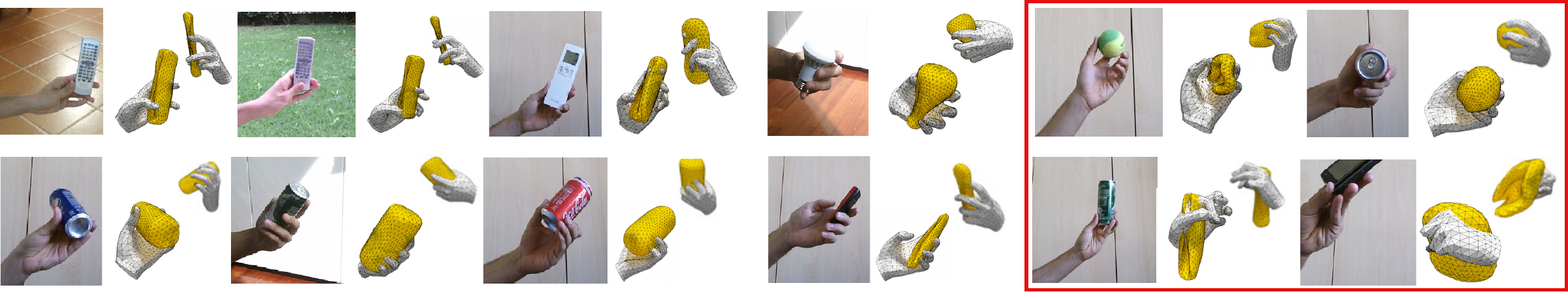}
\mbox{}\vspace{-0.7cm}\\
\caption{Qualitative results on CORe50. Our model, trained only on synthetic data,
  shows robustness
to various hand poses, objects and scenes.
Global hand pose and object outline are well estimated
while fine details are missed.
We present failure cases in the red box.}
\label{fig:core50quali}
\mbox{}\vspace{-1.25cm}\\
\end{figure*}

In Figure~\ref{fig:contactcurve}, we study the effect of introducing our contact loss as a fine-tuning step.
We linearly interpolate $\lambda_{R}$ in [\![0, 1]\!] to explore various
relative weightings of the attraction and repulsion terms.

We find that using $\mathcal{L}_{R}$ in isolation efficiently minimizes the maximum
penetration depth, reducing it by $33\%$ for \datasetname and $68\%$ for $\mathrm{FHB}_C$.
This decrease occurs at the expense of the stability of the grasp in simulation.
Symmetrically, $\mathcal{L}_{A}$ stabilizes the grasps in simulation, but
produces more collisions between hands and objects.
We find that equal weighting of both terms ($\mathcal{L}_R=0.5$) improves $\emph{both}$
physical measures without negatively affecting the reconstruction
metrics on both the synthetic and the real
datasets, as is shown in Table~\ref{table:contact} (last row).
For $\textrm{FHB}_C$, for each metric we report the means and standard deviations for $10$ random seeds.

We find that on the synthetic dataset, decreased penetration is
systematically traded for simulation instability whereas for $\mathrm{FHB}_C$
increasing $\lambda_{R}$ from $0$ to $0.5$ decreases depth penetration
\emph{without} affecting the simulation stability.
Furthermore, for $\lambda_R=0.5$, we observe significant qualitative improvements on
$\mathrm{FHB}_c$ as seen in Figure~\ref{fig:contact_fhb_qual}.

\subsection{Synthetic to real transfer} \label{subsec:synth2real}

\begin{figure}
    \includegraphics[width=.45\linewidth]{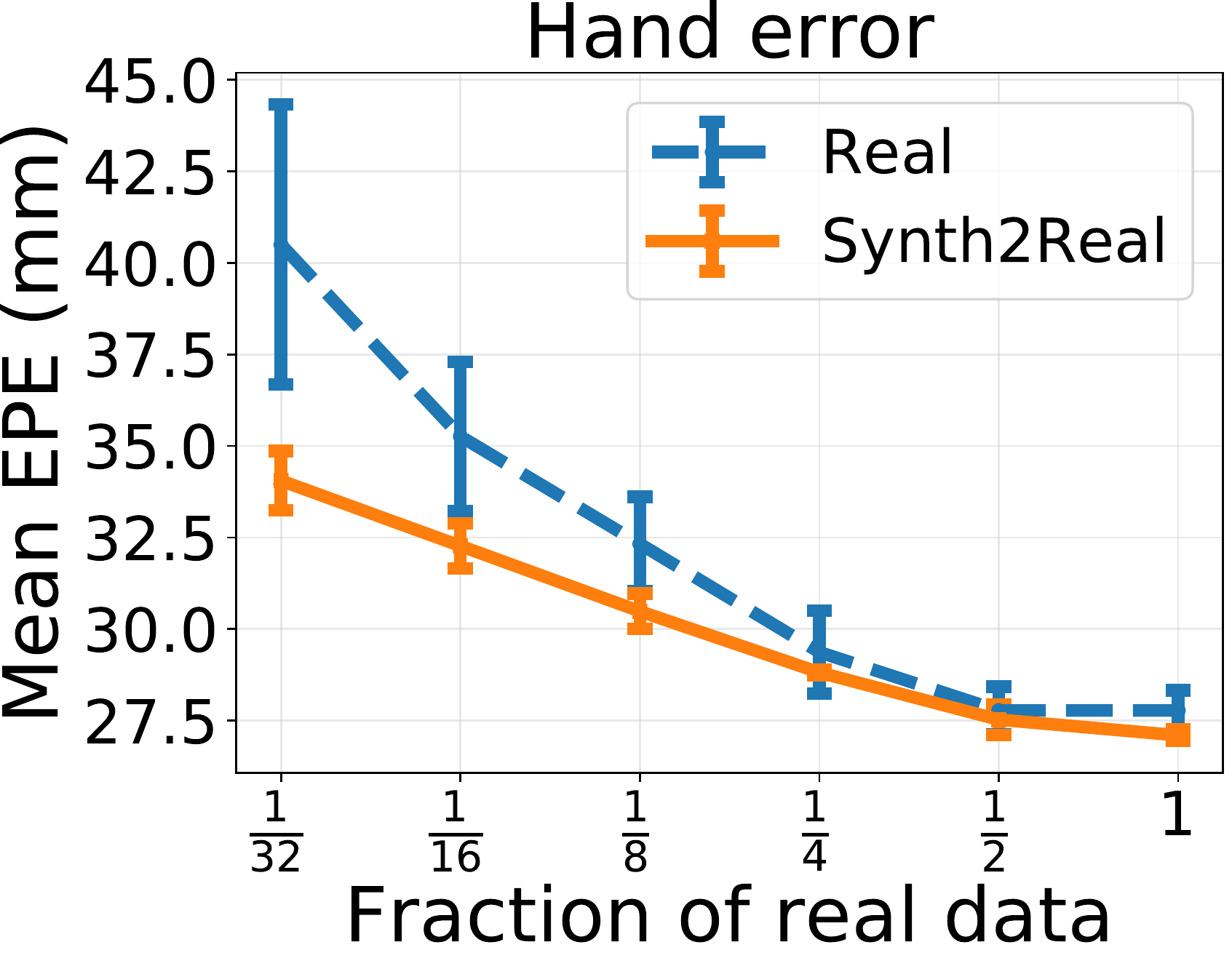}
    \includegraphics[width=.48\linewidth]{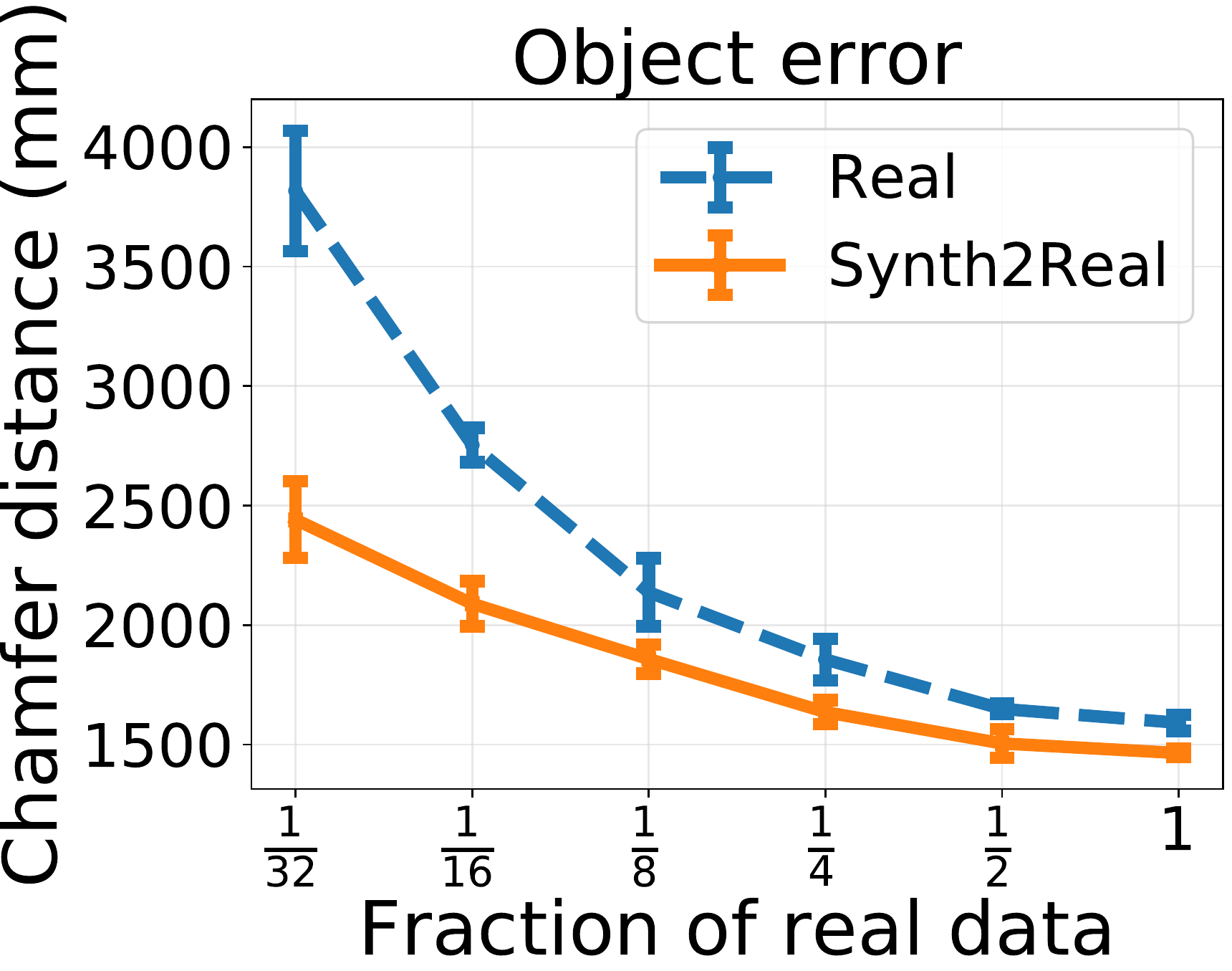}
    \mbox{}\vspace{-.5cm}\\
    \caption{We compare training on FHB only (Real) and pre-training on
        synthetic, followed by fine-tuning on FHB (Synth2Real). As the amount of real data
    decreases, the benefit of pre-training increases. For both the object and the hand
    reconstruction, synthetic pre-training is critical in low-data regimes.}
\label{fig:finetunecomparisonfhb}
\mbox{}\vspace{-1.2cm}\\
\end{figure}

Large-scale synthetic data can be used to pre-train models
in the absence of suitable real datasets.
We investigate the advantages of pre-training on \datasetname when targeting \fhb
and HIC.
We investigate the effect of scarcity of real data on 
\fhb by comparing pairs of networks trained using subsets of the real dataset.
One is pre-trained on ObMan while the other is initialized randomly,
with the exception of the encoders, which are pre-trained on ImageNet~\cite{ILSVRC15}.
For these experiments, we do not add the contact loss and report means
and standard deviations for 5 distinct random seeds.
We find that pre-training on ObMan is beneficial in low data regimes, especially
when less than $1000$ images from the real dataset are used for fine-tuning, see
Figure~\ref{fig:finetunecomparisonfhb}.

\begin{figure}
    \centering
    \includegraphics[width=.99\linewidth]{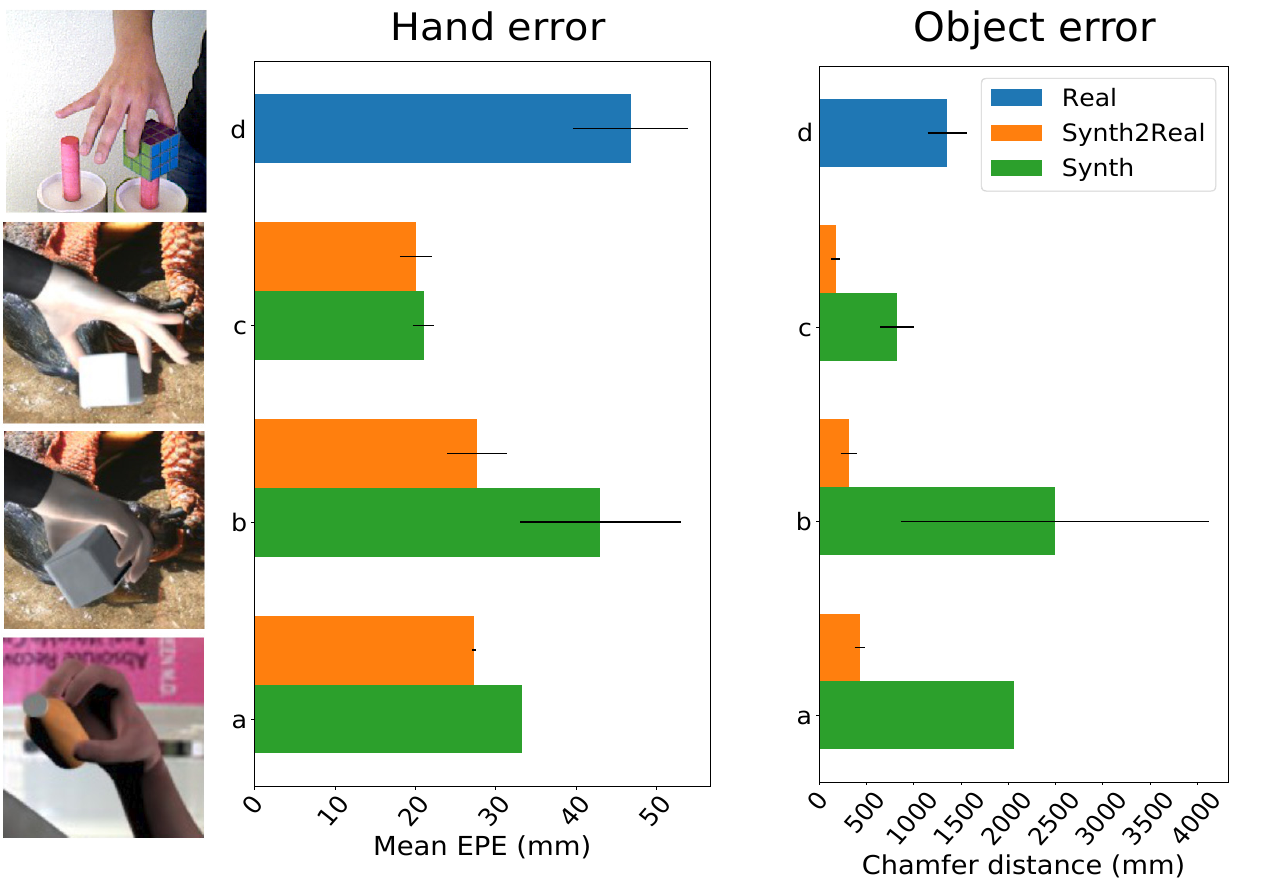}
    \mbox{}\vspace{-.3cm}\\
    \caption{We compare the effect of training with and without fine-tuning on variants of our
    synthetic dataset on HIC. We illustrate each
    dataset (a, b, c, d) with an image sample, see text for definitions.
    Synthetic pre-training, whether or not the target distribution is
    matched, is always beneficial.
}
\label{fig:finetunecomparisonhic}
\mbox{}\vspace{-.8cm}\\
\end{figure}

The HIC training set consists of only $250$ images.
We experiment with pre-training on variants of our synthetic dataset.
In addition to ObMan, to which we refer as (a) in
Figure~\ref{fig:finetunecomparisonhic}, we render $20K$ images for two
additional synthetic
datasets, (b) and (c), which leverage information from the training split of
HIC (d).
We create (b) using our grasping tool to generate automatic
grasps for each of the object models of HIC 
and (c) using the object and pose distributions from the training split of HIC.
This allows to study the importance of sampling hand-object poses from the
target distribution of the real data.
We explore training on (a), (b), (c) with and without fine-tuning on HIC.
We find that pre-training on all three datasets is beneficial for hand and
object reconstructions. The best performance is obtained when pre-training on (c).
In that setup, object performance outperforms training only on
real images even \emph{before} fine-tuning, and significantly improves upon the baseline after.
Hand pose error saturates after the pre-training step, leaving no room
for improvement using the real data.
These results show that when training on synthetic data,
similarity to the target real hand and pose distribution is critical.

\subsection{Qualitative results on CORe50} \label{subsec:core50}

\fhb is a dataset with limited backgrounds, visible magnetic sensors and a very
limited number of subjects and objects.
In this section, we verify the ability of our model trained on \datasetname to
generalize to real data \emph{without} fine-tuning.
CORe50~\cite{pmlr-v78-lomonaco17a} is a dataset which contains hand-object
interactions with an emphasis on the variability of objects and backgrounds.
However no \threeD hand or object annotation is available.
We therefore present qualitative results on this dataset.
Figure~\ref{fig:core50quali} shows that our model generalizes across different
object categories, including \textit{light-bulb},
which does not belong to the categories our model was trained on.
The global outline is well recovered in the camera view
while larger mistakes occur in the perpendicular direction.
More results can be found in Appendix~\ref{app:sec:corequali}.

\section{Conclusions}
\label{sec:conclusions}

We presented an end-to-end approach for joint reconstruction
of hands and objects given a single RGB image as input.
We proposed a novel contact loss that enforces
physical constraints on the interaction between the
two meshes. Our results and the ObMan dataset
open up new possibilities for research on
modeling object manipulations. Future directions
include learning grasping affordances from
large-scale
visual data, and
recognizing complex and dynamic hand actions.

\mbox{}\vspace{-.75cm}\\

\paragraph{Acknowledgments.}
\footnotesize
This work was supported in part by ERC grants ACTIVIA and ALLEGRO,  the MSR-Inria
joint lab, the Louis Vuitton ENS Chair on AI and the DGA
project DRAAF.  
We thank Tsvetelina Alexiadis, Jorge Marquez and Senya Polikovsky
from MPI for help with scan acquisition, Joachim Tesch for the hand-object
rendering, Mathieu Aubry and Thibault Groueix for advices on AtlasNet,
David Fouhey for feedback.
MJB has received research gift funds from Intel, Nvidia,
Adobe, Facebook, and Amazon. While MJB is a part-time employee of Amazon, his
research was performed solely at, and funded solely by, MPI.
MJB has financial interests in Amazon and Meshcapade GmbH.

{\small
\balance
\bibliographystyle{ieee}
\bibliography{references}
}

{\normalsize
\clearpage
\renewcommand{\thefigure}{A.\arabic{figure}} 
\setcounter{figure}{0} 
\renewcommand{\thetable}{A.\arabic{table}}
\setcounter{table}{0} 

\appendix
\section*{APPENDIX}

Our main paper proposed a method for joint reconstruction of hands and objects.
Below we present complementary analysis for hand-only reconstruction in
Section~\ref{app:sec:mano} and object-only reconstruction in Section~\ref{app:sec:objrec}.
Section~\ref{app:sec:impdetails} presents implementation details.

\section{Hand pose estimation}
\label{app:sec:mano}
We first present an ablation study for the different losses we defined on the MANO
 hand model (Section~\ref{app:subsec:manoloss}).
Then, we study the latent hand representation
(Section~\ref{app:subsec:manorepresentation}).
Finally, we validate our hand pose estimation branch and demonstrate its
competitive performance compared to the state-of-the-art methods on a benchmark
dataset (Section~\ref{app:subsec:handsoa}).

\subsection{Loss study on MANO}
\label{app:subsec:manoloss}

As explained in Section~\ref{subsec:hand} of the main paper,
we define three losses for the differentiable hand model
while training our network: (i) vertex positions $\mathcal{L}_{V_{\mathit{Hand}}}$,
(ii) joint positions $\mathcal{L}_J$, and
(iii) shape regularization $\mathcal{L}_\beta$.
The shape is only predicted in the presence of $\mathcal{L}_\beta$.
In the absence of shape regularization, when only sparse keypoint supervision is
provided, predicting $\beta$ without regularizing it produces extreme
deformations of the hand mesh, and we therefore fix $\beta$ to the average hand
shape.

Table~\ref{table:manoloss} summarizes the contribution of each
of these losses. Note that the dense vertex supervision is 
available on our synthetic dataset ObMan, and not available
on the real datasets FHB~\cite{hernando2018cvpr} and
StereoHands~\cite{stereohands2016}.

\begin{table}[h]
    \centering
    \resizebox{.9\linewidth}{!}{
        \begin{tabular}{lccc}
            \toprule
         &  ObMan & FHB & StereoHands \\
            \midrule
        $\mathcal{L}_{J}$ & 13.5 & 28.1 & 11.4 \\
        $\mathcal{L}_{J} + \mathcal{L}_{\beta}$ & 11.7 & \textbf{26.5} &
        \textbf{10.0} \\
        $\mathcal{L}_{V_{\mathit{Hand}}}$ & 14.0 & - & - \\
        $\mathcal{L}_{V_{\mathit{Hand}}} + \mathcal{L}_{\beta}$ & 12.0 & - & - \\
        $\mathcal{L}_{V_{\mathit{Hand}}} + \mathcal{L}_{J} + \mathcal{L}_{\beta}$ &
        \textbf{11.6} & - & - \\
            \bottomrule
        \end{tabular}
    }
    \caption{We report the mean end-point error (mm) to study
    different losses defined on MANO.
    We experiment with the loss on
    3D vertices ($\mathcal{L}_{V_{\mathit{Hand}}}$),
    3D joints ($\mathcal{L}_{J}$),
    and shape regularization ($\mathcal{L}_{\beta}$).
    We show the results of training and testing on our synthetic ObMan dataset,
    as well as the real datasets FHB~\cite{hernando2018cvpr}
    and StereoHands~\cite{stereohands2016}.}
    \label{table:manoloss}
\end{table}

We find that predicting $\beta$ while regularizing it with $\mathcal{L}_{\beta}$
significantly improves the mean end-point-error on keypoints. On the synthetic
dataset ObMan, we find that adding $\mathcal{L}_{V}$ yields a small
additional improvement.
We therefore use all three losses whenever dense vertex supervision is
available, and $\mathcal{L}_{J}$ in conjunction with $\mathcal{L}_{\beta}$ when
only keypoint supervision is provided.

\subsection{MANO pose representation}
\label{app:subsec:manorepresentation}
As described in Section~\ref{subsec:hand} of the main paper,
our hand branch outputs a 30-dimensional vector
to represent the hand. These are the 30 first PCA
components from the 45-dimensional full pose space.
We experiment with different dimensionality for the latent
hand representation and summarize our findings in Table~\ref{table:manopca}.
While low-dimensionality fails to capture
some poses present in the datasets, we do not
observe improvements after increasing the dimensionality
more than 30. Therefore, we use this value for all experiments
in the main paper.

\begin{table}[h]
    \centering
    \resizebox{.8\linewidth}{!}{
        \begin{tabular}{lccccc}
            \toprule
        \#PCA comps. & 6 & 15 & 30 & 45 \\
            \midrule
        FHB & 28.2 & 27.5 & \textbf{26.5} & 26.9 \\
        StereoHands & 13.9 & 11.1 & \textbf{10.0} & \textbf{10.0} \\
        ObMan & 23.4 & 13.3 & 11.6 & \textbf{11.2} & \\
            \bottomrule
        \end{tabular}
    }
    \caption{We report the mean end-point error on error on multiple
    datasets to
    study the effect of the number of PCA hand pose components
    for the latent MANO representation.}
    \label{table:manopca}
\end{table}

\subsection{Comparison with the state of the art}
\label{app:subsec:handsoa}

\begin{figure}[h]
    \centering
    \includegraphics[width=.32\linewidth]{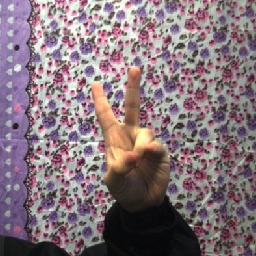}
    \includegraphics[width=.32\linewidth]{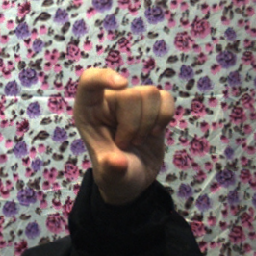}
    \includegraphics[width=.32\linewidth]{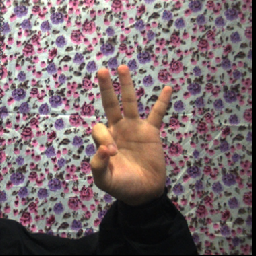}
    \includegraphics[trim={0 4cm 0 0},clip, width=.32\linewidth]{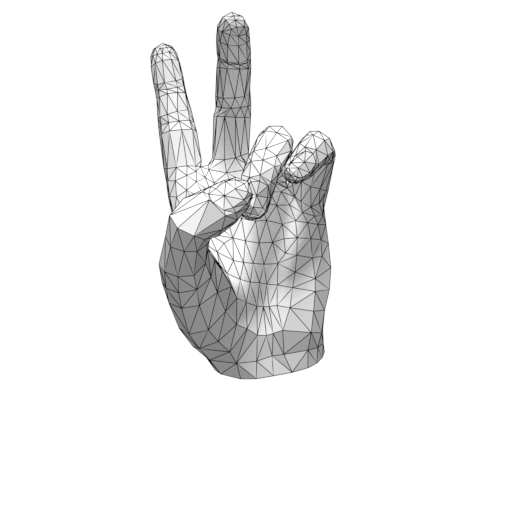}
    \includegraphics[trim={0 4cm 0 0},clip, width=.32\linewidth]{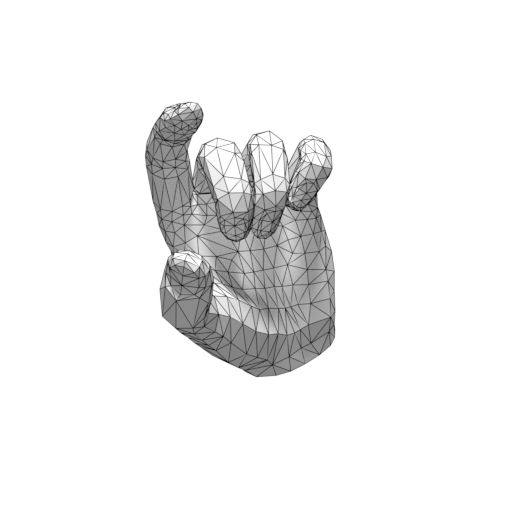}
    \includegraphics[trim={0 4cm 0 0},clip, width=.32\linewidth]{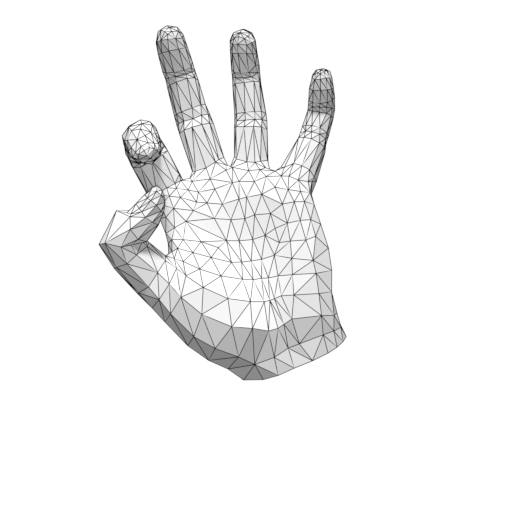}
    \caption{Qualitative results on the test sequence of the StereoHands dataset.}
\label{fig:stereohands}
\end{figure}
\begin{figure}
\begin{center}
    \includegraphics[trim={0 0 0 1.5cm},clip, width=0.6\linewidth]{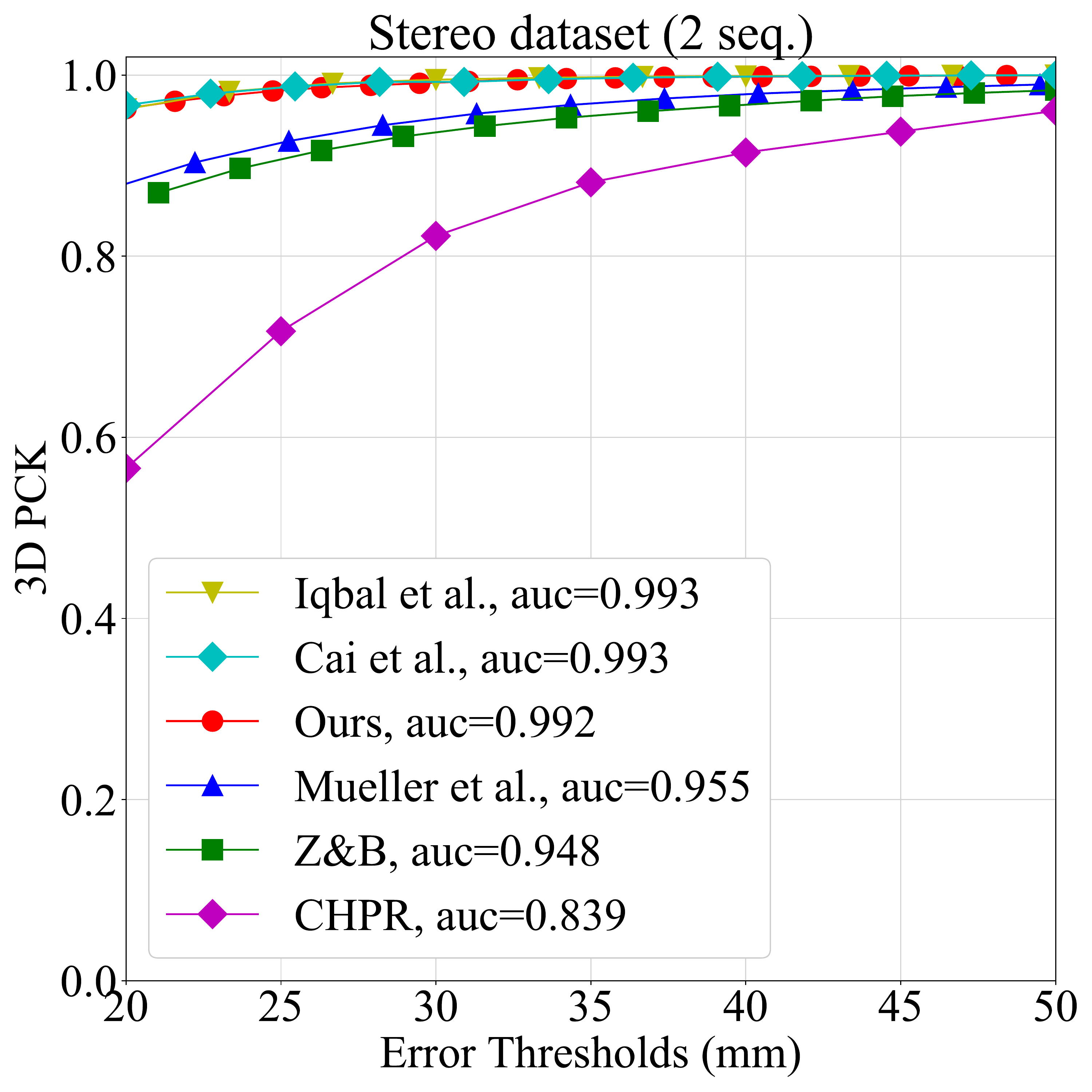}
\end{center}
   \caption{
   We compare our root-relative 3D hand pose estimation on Stereohands
   to the state-of-the-art methods from
   Iqbal~\etal~\cite{iqbal2018ECCV},
   Cai~\etal~\cite{cai2018_weakly}, Mueller~\etal~\cite{GANeratedHands_CVPR2018},
   Zimmermann and Brox~\cite{brox:ICCV:2017}, and CHPR~\cite{Sun2015CascadedHP}.
   }
\label{fig:handsoa}
\end{figure}

Using the MANO branch of the network, we can also estimate the hand pose
for images in which the hands are not interacting with objects, and compare our
results with previous methods.
We train and test on the StereoHands dataset~\cite{stereohands2016}, and follow
the evaluation protocol
of~\cite{iqbal2018ECCV,brox:ICCV:2017,GANeratedHands_CVPR2018} by training on 10
sequences from StereoHands and testing on the 2 remaining ones.
For fair comparison, we add a palm joint to the MANO model by averaging the positions of two vertices on the front and back of the
hand model at the level of the palm.
Although the hand shape parameter $\beta$ allows to capture the variability of hand shapes
which occurs naturally in human populations, it does not account for the
discrepancy between different joint conventions.
To account for skeleton mismatch, we add a linear layer initialized
to identity which maps from the MANO joints to the final joint annotations.

We  report the area under the curve (auc)
on the percentage of correct keypoints (PCK).
Figure~\ref{fig:handsoa} shows that our differentiable hand model
is on par with the state of the art.
Note that the StereoHands benchmark is close to saturation. In contrast to other
methods~\cite{cai2018_weakly,iqbal2018ECCV,GANeratedHands_CVPR2018,brox:ICCV:2017,Sun2015CascadedHP}
that only predicts sparse skeleton keypoints,
our model produces a \textit{dense} hand mesh.
Figure~\ref{fig:stereohands} presents some qualitative results
from this dataset.

\section{Object reconstruction}
\label{app:sec:objrec}
In the following, we validate our design choices
for the object reconstruction branch.
We experiment with object reconstruction
(i) in the camera viewpoint (Section~\ref{app:subsec:objectcanonical}) and
(ii) with regularization losses (Section~\ref{app:subsec:objectregularization}).

\subsection{Canonical versus camera view reconstruction}
\label{app:subsec:objectcanonical}

As explained in Section~\ref{subsec:object} of the main paper,
we perform object reconstructions in the camera coordinate frame.
To validate that AtlasNet~\cite{groueix2018} can successfully
predict objects in camera view as well as in canonical view,
we reproduce the training setting of the original paper~\cite{groueix2018}.
We use the setting where 2500 points are sampled on a sphere
and train on the rendered images from ShapeNet~\cite{choy20163d}.
To obtain the rotated reference for the object, we apply the ground truth
azimuth and elevation provided with the renderings so that the 3D
ground truth matches the camera view.
We use the original hyperparameters (Adam~\cite{adam2014} with a
learning rate of 0.001) and train both networks for 25 epochs.
Both for supervision and evaluation metrics, we report the Chamfer distance
$\mathcal{L}_{V_{Obj}} = \frac{1}{2} (\sum_{p} min_q \Vert p - q \Vert^2_2 +
\sum_{q} min_p \Vert q - p \Vert^2_2)$ where $q$ spans the predicted vertices
and $p$ spans points uniformly sampled on the surface of the ground truth object.
We always sample the same number of points on the surface as there are vertices
in the predicted mesh.
We find that both numerically and qualitatively the performance is
comparable for the two settings.
Some reconstructed meshes in camera view are shown in Figure~\ref{fig:atlasnetcanonical}.
For better readability they also multiply the Chamfer loss by $1000$. 
In order to provide results directly comparable with the original
paper~\cite{groueix2018}, we also report numbers with the same scaling in Table~\ref{table:objectreconstruction}.
Table~\ref{table:objectreconstruction} reports the Chamfer distances for their
released model, our reimplementation in canonical view, and our implementation
in non-canonical view.
We find that our implementation allows us to train a model with similar
performances to the released model.
We observe no numerical or qualitative loss in performance
when predicting the camera view instead of the canonical one.

\begin{figure}[h]
    \centering
        \includegraphics[width=0.32\linewidth]{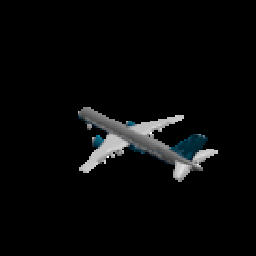}
        \includegraphics[width=0.32\linewidth]{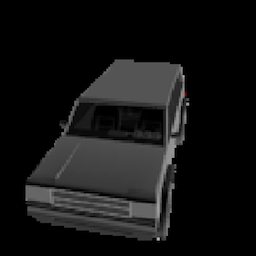}
        \includegraphics[width=0.32\linewidth]{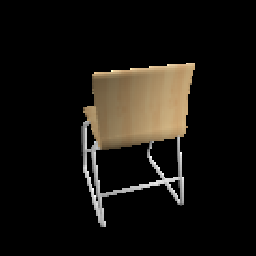}
        \includegraphics[width=0.32\linewidth]{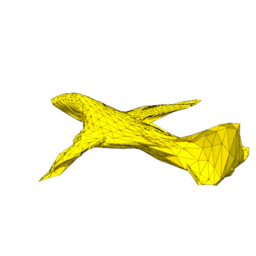}
        \includegraphics[width=0.32\linewidth]{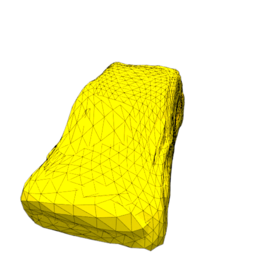}
        \includegraphics[width=0.32\linewidth]{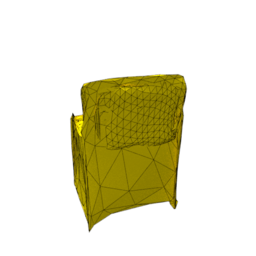}
    \caption{Renderings from ShapeNet models and our
    corresponding reconstructions in camera view.}
\label{fig:atlasnetcanonical}
\end{figure}
\begin{table}[h]
  \centering
  \begin{tabular}{lc}
    \toprule
    & Object error\\
    \midrule
    Canonical view~\cite{groueix2018} & 4.87 \\
    Canonical view (ours) & 4.88 \\
    Camera view (ours) & 4.88 \\
    \bottomrule
  \end{tabular}
  \caption{Chamfer loss ($\times 1000$) for 2500 points in
    the canonical view and camera view show no degradation
    from predicting the camera view reconstruction. We compare
    our re-implementation
    to the results
    provided by~\cite{groueix2018} on their code page
    \href{https://github.com/ThibaultGROUEIX/AtlasNet}{https://github.com/ThibaultGROUEIX/AtlasNet}.
  }
  \label{table:objectreconstruction}
\end{table}

\subsection{Object mesh regularization}
\label{app:subsec:objectregularization}

We find that in the absence of explicit regularization on their quality, the
predicted meshes can be very irregular.  Sharp discontinuities in curvature
occur in regions where the ground truth mesh is smooth, and the mesh triangles
can be of very different dimensions.  These shortcomings can be observed on
all three reconstructions in~Figure~\ref{fig:atlasnetcanonical}.
Following recent work on mesh estimation from image
inputs~\cite{wang2018pixel2mesh,cmrKanazawa18,groueix2018b}, we introduce
regularization terms on the object mesh.

\noindent\textbf{Laplacian smoothness regularization ($\mathcal{L}_{L}$).} In
order to avoid unwanted discontinuities in the curvature of the mesh, we enforce
a local prior of smoothness. We use the discrete Laplace-Beltrami operator to
estimate the curvature at each mesh vertex position, as we have no prior on the
final shape of the geometry, we compute the graph laplacian $L$ on our mesh, which only
takes into account adjacency between mesh vertices. Multiplying the laplacian
$L$ by the positions of the object vertices $\mathcal{V}_{Obj}$ produces vectors
which have the same direction as the vertex normals and their norm proportional
to the curvature.
Minimizing the norm of these vector therefore minimizes the curvature.
We minimize the mean curvature over all vertices in order to encourage
smoothness on the mesh.

\noindent\textbf{Laplacian edge length regularization ($\mathcal{L}_{E}$).} 
$\mathcal{L}_{E}$ penalizes configurations in which the edges of the mesh
 have different lengths. 
 The edge regularization is defined as:

 \begin{equation}
     \mathcal{L}_E = \frac{1}{|\mathcal{E}_L|} \sum_{l \in \mathcal{E}_L} |l^2
     - \mu({\mathcal{E}_L^2})|,
 \end{equation}
where $\mathcal{E}_L$ is the set of edge lengths, defined as the L2 norms of
the edges, and $\mu({\mathcal{E}_L^2)}$ is the average of the square of edge
lengths.

To evaluate the effect of the two regularization terms we train four different
models. We train a model without any regularization, two models for which only
one of the two regularization terms are active, and finally a model for
which the two regularization terms are applied simultaneously.
Each of these models is trained for $200$ epochs.

Figure~\ref{fig:objectregularization} shows the qualitative benefits of
each term.
While edge regularization $\mathcal{L}_{E}$ alone already significantly improves the quality of
the predicted mesh, note that unwanted bendings of the mesh still occur, for
instance in the last row for the cellphone reconstruction. Adding the laplacian
smoothness $\mathcal{L}_{L}$ resolves these irregularities.
However, adding each regularization term negatively affects the final
reconstruction score. Particularly we observe that introducing edge
regularization increases the Chamfer loss by 22\% while significantly improving
the perceptual quality of the predicted mesh.
Introducing the regularization terms contributes to the coarseness of
the object reconstructions, as can be observed on the third row, where sharp
curvatures of the object in the input image are not captured in the reconstruction.

\begin{figure}
    \centering
    \resizebox{\linewidth}{!}{
    \begin{tabular}{ccccc}
    \includegraphics[width=.32\linewidth]{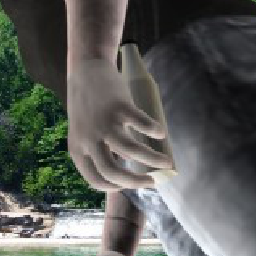}  &
    \includegraphics[trim={\trimlen cm \trimlen cm \trimlen cm \trimlen cm}, clip, width=.32\linewidth]{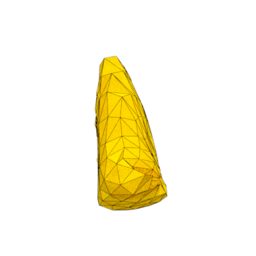} &
    \includegraphics[trim={\trimlen cm \trimlen cm \trimlen cm \trimlen cm}, clip, width=.32\linewidth]{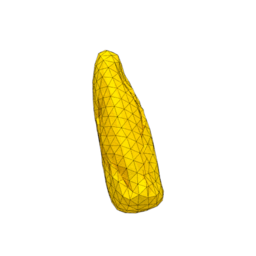} &
    \includegraphics[trim={\trimlen cm \trimlen cm \trimlen cm \trimlen cm}, clip, width=.32\linewidth]{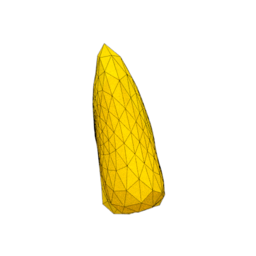} &
    \includegraphics[trim={\trimlen cm \trimlen cm \trimlen cm \trimlen cm}, clip, width=.32\linewidth]{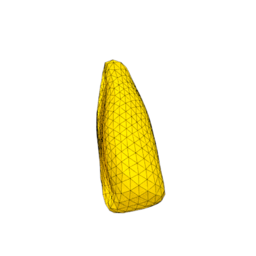} \\ 

    \includegraphics[width=.32\linewidth]{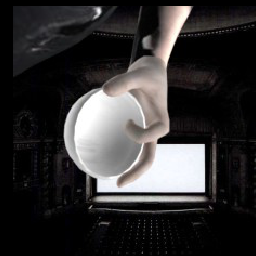}  &
    \includegraphics[trim={\trimlen cm \trimlen cm \trimlen cm \trimlen cm}, clip, width=.32\linewidth]{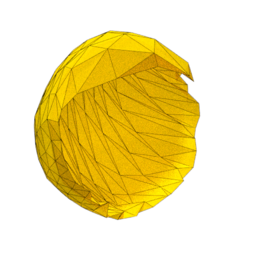} &
    \includegraphics[trim={\trimlen cm \trimlen cm \trimlen cm \trimlen cm}, clip, width=.32\linewidth]{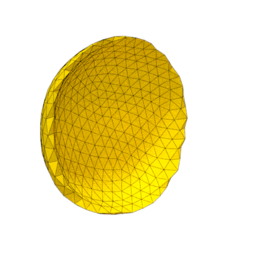} &
    \includegraphics[trim={\trimlen cm \trimlen cm \trimlen cm \trimlen cm}, clip, width=.32\linewidth]{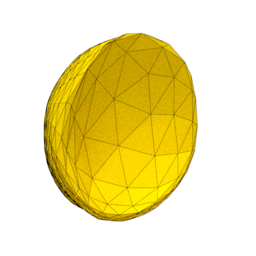} &
    \includegraphics[trim={\trimlen cm \trimlen cm \trimlen cm \trimlen cm}, clip, width=.32\linewidth]{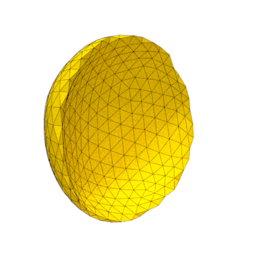} \\ 

    \includegraphics[width=.32\linewidth]{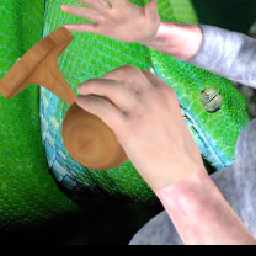}  &
    \includegraphics[trim={\trimlen cm \trimlen cm \trimlen cm \trimlen cm}, clip, width=.32\linewidth]{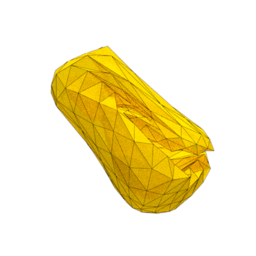} &
    \includegraphics[trim={\trimlen cm \trimlen cm \trimlen cm \trimlen cm}, clip, width=.32\linewidth]{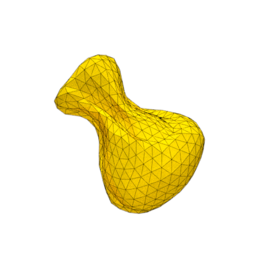} &
    \includegraphics[trim={\trimlen cm \trimlen cm \trimlen cm \trimlen cm}, clip, width=.32\linewidth]{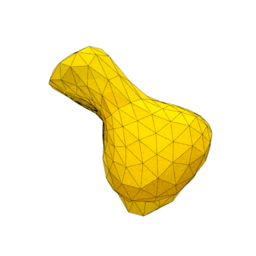} &
    \includegraphics[trim={\trimlen cm \trimlen cm \trimlen cm \trimlen cm}, clip, width=.32\linewidth]{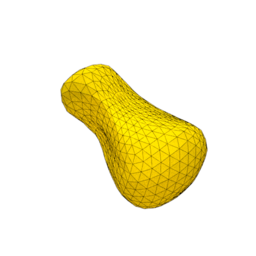} \\ 

     \includegraphics[width=.32\linewidth]{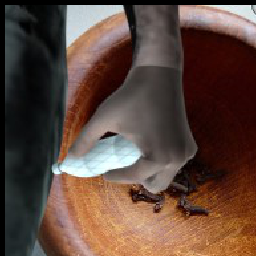}  &
     \includegraphics[clip, width=.32\linewidth]{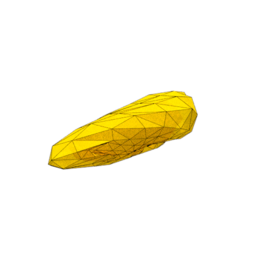} &
    \includegraphics[trim={\trimlen cm \trimlen cm \trimlen cm \trimlen cm}, clip, width=.32\linewidth]{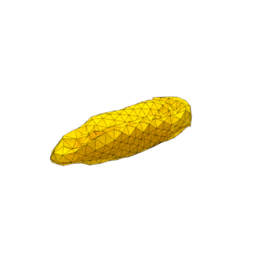} &
    \includegraphics[trim={\trimlen cm \trimlen cm \trimlen cm \trimlen cm}, clip, width=.32\linewidth]{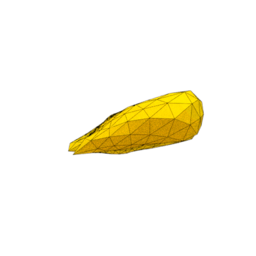} & \includegraphics[trim={\trimlen cm \trimlen cm \trimlen cm \trimlen cm}, clip, width=.32\linewidth]{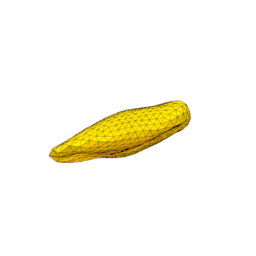} \\ 
    \includegraphics[width=.32\linewidth]{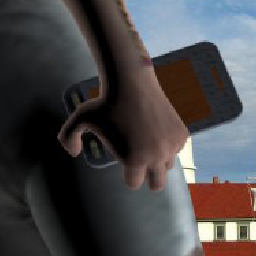}  &
     \includegraphics[trim={\trimlen cm \trimlen cm \trimlen cm \trimlen cm}, clip, width=.32\linewidth]{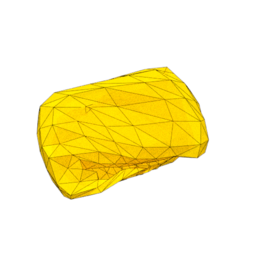} &
    \includegraphics[trim={\trimlen cm \trimlen cm \trimlen cm \trimlen cm}, clip, width=.32\linewidth]{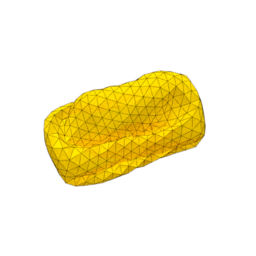} & \includegraphics[trim={\trimlen cm \trimlen cm \trimlen cm \trimlen cm}, clip, width=.32\linewidth]{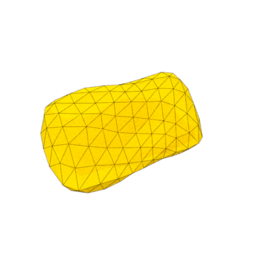} & \includegraphics[trim={\trimlen cm \trimlen cm \trimlen cm \trimlen cm}, clip, width=.32\linewidth]{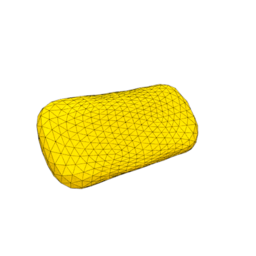} \\ \toprule
    & \LARGE No reg. & \LARGE $\mathcal{L}_E$ & \LARGE $\mathcal{L}_L$ & \LARGE $\mathcal{L}_E +
    \mathcal{L}_L$ \\ \midrule
     \LARGE Object error &  \LARGE 0.0246 &  \LARGE 0.0286 &  \LARGE 0.0258 &  \LARGE  0.0292 \\ \bottomrule
    \end{tabular}
    }
    \caption{We show the benefits from each term of the regularization.
        Using both the $\mathcal{L}_E$ and $\mathcal{L}_L$ in conjunction
        improves the visual quality of the predicted triangulation while preserving the
shape of the object.}
\label{fig:objectregularization}
\end{figure}

\section{Implementation details}
\label{app:sec:impdetails}
We give implementation details on our training procedure
(Section~\ref{app:subsec:training}) and our automatic
grasp generation (Section~\ref{app:subsec:grasps}).

\subsection{Training details}
\label{app:subsec:training}

For all our experiments,
we use the Adam optimizer~\cite{adam2014}.
As we observe instabilities in validation curves when training on synthetic
datasets,
we freeze the batch normalization layers.
This fixes their weights to the original values
from the ImageNet~\cite{ILSVRC15} pre-trained ResNet18~\cite{He2015}.

For the final model trained on ObMan, we first train the (normalized) object
branch using $\mathcal{L}^{n}_{\mathit{Object}}$ for 250
epochs, we start with a learning rate of $10^{-4}$ and decrease it to $10^{-5}$
at epoch 200.
We then freeze the object encoder and the AtlasNet decoder,
as explained in Section~\ref{subsec:object} of the main paper.
We further train the full network with $\mathcal{L}_{\mathit{Hand}}+\mathcal{L}_{\mathit{Object}}$ for $350$ additional epochs, decreasing the
learning rate from $10^{-4}$ to $10^{-5}$ after the first $200$ epochs.

When fine-tuning from our main model trained on synthetic data to smaller real datasets,
we unfreeze the object reconstruction branch.

For the FHB$_c$ dataset, we train all the parts of the network simultaneously with
the supervision $\mathcal{L}_{\mathit{Hand}}+\mathcal{L}_{\mathit{Object}}$ for $400$ epochs, decreasing the learning
rate from $10^{-4}$ to $10^{-5}$ at epoch $300$.

When fine-tuning our models with the additional contact loss,
$\mathcal{L}_{\mathit{Hand}}+\mathcal{L}_{\mathit{Object}}+\mu_C \mathcal{L}_{\mathit{Contact}}$, we
use a learning rate of $10^{-5}$.
We additionally set the momentum of the Adam optimizer~\cite{adam2014} to zero,
as we find that momentum affects negatively the training stability 
when we include the contact loss.

In all experiments, we keep the relative weights between different losses
as provided in the main paper and normalize them so
that the sum of all the weights equals 1.

\subsection{Heuristic metric for sorting GraspIt grasps}
\label{app:subsec:grasps}

\begin{figure*}
\centering
\includegraphics[width=0.99\linewidth]{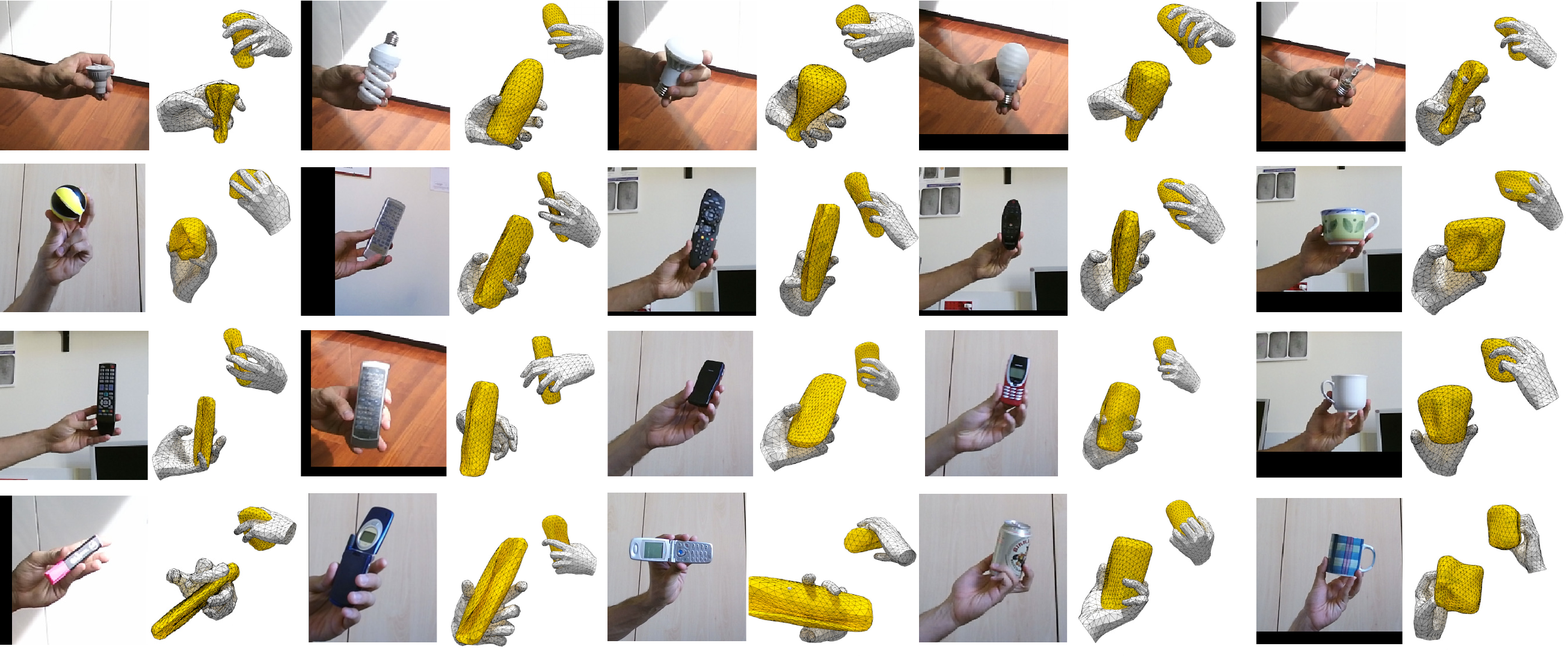}
\mbox{}\vspace{-0.4cm}\\
\caption{
    Qualitative results on CORe50 dataset. We present additional hand-object reconstructions for a
    variety of object categories and object instances, spanning various hand
    poses and object shapes.
}
\mbox{}\vspace{-0.4cm}\\
\label{fig:core50picked}
\end{figure*}

We use GraspIt~\cite{Miller2004} to generate grasps for the ShapeNet object
models. GraspIt generates a large variety of grasps by exploring different
initial hand poses. However, some initializations do not produce good grasps.
Similarly to~\cite{GraspDatabase2009} we filter the grasps in a post-processing
step in order to retain grasps of good quality according to a heuristic metric
we engineer for this purpose.

For each grasp, GraspIt provides two grasp quality metrics $\varepsilon$ and
$v$ ~\cite{Ferrari1992PlanningOG}.
Each grasp produced by GraspIt~\cite{Miller2004} defines contact points between
the hand and the object. Assuming rigid contacts with friction, we can compute
the space of wrenches which can be resisted by the grasp: the grasp wrench
space (GWS).
This space is normalized with relation to the scale of the object, defined as
the maximum radius of the object, centered at its center of mass.
The grasp is suitable for any task that involves external wrenches that lie
within the GWS.
$v$ is the volume of the 6-dimensional GWS, which quantifies the range of
wrenches the grasp can resist.
The GWS can further be characterized by the radius $\varepsilon$
of the largest ball which is centered at the origin and inscribed in the grasp
wrench space.  $\varepsilon$ is the maximal wrench norm that can be balanced by
the contacts for external wrenches applied coming from arbitrary directions.
$\varepsilon$  belongs to $[0, 1]$ in the scale-normalized GWS,
and higher values are associated with a higher robustness to external wrenches.

We require a single value to reflect the quality of the grasp in order
to sort different grasps.
We use the norm of the $[\varepsilon, v]$ vector in our heuristic measure of
grasp quality. 
We find that in the grasps produced by GraspIt, power grasps, as defined
by~\cite{romero:THMS:2016} in which larger surfaces of the hand and the object
are in contact, are rarely produced.
To allow for a larger proportion of power grasps, we use a multiplier
$\gamma_{palm}$ which we empirically set to $1$ if the palm is not in
contact and $3$ otherwise. 
We further favor grasps in which a large number of phalanges are in contact with
the object by weighting the final grasp score using $N_{p}$, the number of
phalanges in contact with the object, which is computed by the software.

The final grasp quality score $G$ is defined as:
\begin{equation} G = \gamma_{palm} \sqrt{N_{p}} \Vert \varepsilon, v \Vert_{2}.
 \end{equation}
We find that keeping the two best grasps for each object
produces both diverse grasps and grasps of good quality.

\section{Qualitative results on CORe50 dataset} \label{app:sec:corequali}

We present additional qualitative results on the CORe50~\cite{pmlr-v78-lomonaco17a} dataset.
We present a variety of diverse input images from CORe50 in Figure~\ref{fig:core50picked}
alongside the predictions of our final model trained solely on ObMan.

The first row presents results on various shapes of light bulbs. Note that this
category is not included in the synthetic object models of ObMan.
Our model can therefore generalize across object categories.
The last column shows some reconstructions of mugs, showcasing the topological
limitations of the sphere baseline of AtlasNet which cannot, by construction,
capture handles.

However, we observe that the object shapes are often coarse, and
that fine details such as phone antennas are not reconstructed.
We also observe errors in the relative position between the object and the hand,
which is biased towards predicting the object's centroid in the
palmar region of the hand, see Figure~\ref{fig:core50picked}, fourth column.
As hard constraints on collision are not imposed, hand-object
interpenetration occurs in some configurations, for instance in the top-right
example.
In the bottom-left example we present a failure case where the hand
pose violates anatomical constraints. Note that while our model predicts hand
pose in a low-dimensional space, which implicitly regularizes hand poses,
anatomical validity is not guaranteed.



}

\end{document}